\documentclass[review,12pt,3p]{elsarticle}
\usepackage{amssymb}
\usepackage{amsmath}
\usepackage{graphicx}
\usepackage{multirow}
\usepackage{amsmath,amssymb,amsfonts}
\usepackage{hyperref} 

\usepackage{bm}
\usepackage{url}
\usepackage{booktabs}
\usepackage{adjustbox}
\usepackage{array}
\usepackage[linesnumbered,lined,boxed,commentsnumbered,ruled,vlined]{algorithm2e}
\usepackage{siunitx}  % 提供 S 列类型和数字格式化

\usepackage{color}

\journal{Pattern Recognition}

\begin{document}

\begin{frontmatter}



\title{UIFV: Data Reconstruction Attack in Vertical Federated Learning}


% \author{Jirui Yang} %% Author name

% \affiliation{organization={},%Department and Organization

\author[1]{Jirui Yang}
\ead{yangjr23@m.fudan.edu.cn}

\author[5]{Peng Chen}
\ead{chenpenghehedawang@gmail.com}

\author[1,2]{Zhihui Lu}
\ead{lzh@fudan.edu.cn}

\author[4]{Qiang Duan}
\ead{qduan@psu.edu}

\author[1]{Yubing Bao}
\ead{ybbao23@m.fudan.edu.cn}

% \cortext[3]{Corresponding author: Zhihui Lu}

\affiliation[1]{organization={School of Computer Science, Fudan University},
            %addressline={}, 
            city={Shanghai},
%          citysep={}, % Uncomment if no comma needed between city and postcode
            postcode={200433}, 
            %state={},
            country={China}}

\affiliation[2]{organization={Shanghai Blockchain Engineering
Research Center},
            %addressline={}, 
            city={Shanghai},
%          citysep={}, % Uncomment if no comma needed between city and postcode
            postcode={200433}, 
            %state={},
            country={China}}

\affiliation[3]{organization={Institute of Financial Technology,
Fudan University},
            %addressline={}, 
            city={Shanghai},
%          citysep={}, % Uncomment if no comma needed between city and postcode
            postcode={200433}, 
            %state={},
            country={China}}

\affiliation[4]{organization={Information Sciences \& Technology,
Pennsylvania State University},
            %addressline={}, 
            city={PA},
%          citysep={}, % Uncomment if no comma needed between city and postcode
            postcode={16802}, 
            %state={},
            country={USA}}
            
\affiliation[5]{organization={School of Software, Nanjing University of Information Science and Technology},
            %addressline={}, 
            city={Nanjing},
%          citysep={}, % Uncomment if no comma needed between city and postcode
            postcode={210044}, 
            %state={},
            country={China}}

%% Abstract
\begin{abstract}
Vertical Federated Learning (VFL) enables collaborative machine learning without the need for participants to share their raw private data. However, recent studies have uncovered privacy risks, where adversaries might reconstruct sensitive features through data leakage during the learning process. Although existing data reconstruction methods are effective to some extent, they exhibit limitations in VFL scenarios, as initiating an attack requires meeting more stringent conditions. To gain a comprehensive understanding of the risks of data reconstruction in VFL, this paper proposes a unified framework, the Unified InverNet Framework in VFL (UIFV), for data reconstruction under realistic black-box threat models. Within the UIFV framework, we consider four attack scenarios, strictly adhering to VFL protocols to maintain confidentiality. Experiments on four datasets show that our methods significantly outperform state-of-the-art techniques in terms of applicability and attack precision. Our work reveals severe privacy vulnerabilities within VFL systems that pose real threats to practical VFL applications, thus confirming the necessity of further enhancing privacy protection in the VFL architecture. Overall, this paper provides a thorough analysis of the risks of data reconstruction in VFL and offers important guidance to enhance the security of VFL deployments.
\end{abstract}


% \begin{highlights}
% \item A novel method for data reconstruction attack in vertical federated learning (VFL) is proposed. This method leverages intermediate feature data to reconstruct the original data and significantly improves attack accuracy compared to existing data reconstruction techniques.
% \item A versatile UIFV framework has been developed, supporting various black-box attack scenarios and tailored to different adversary capabilities in VFL environments.
% \item Serious privacy risks in VFL systems are revealed across four attack scenarios, highlighting the necessity of enhancing privacy protection. The research results provide useful insights into enhancing VFL security.
% \end{highlights}

%% Keywords
\begin{keyword}
 Vertical federated learning \sep privacy risks \sep data leakage \sep Unified InverNet Framework (UIFV) \sep data reconstruction \sep intermediate feature data.

\end{keyword}

\end{frontmatter}

\section{Introduction}

In today's field of artificial intelligence (AI), the integration of federated learning is regarded as a truly transformative strategy. This unique machine learning approach distributes the model training process across multiple devices and subsequently combines the model updates from these individual devices into a comprehensive global model. This process achieves a delicate balance by protecting data privacy while utilizing large-scale datasets for effective model training and optimization. Federated learning circumvents the need for direct access to or transmission of raw data, thereby significantly reducing the risk of data breaches. At the same time, the model's ability to train on large datasets greatly enhances its performance and generalization capabilities. \textcolor{black}{Now, federated learning is increasingly applied to real-life applications such as mobile keyboard prediction\cite{hard2018federated}, healthcare\cite{guan2024federated}, and purchase recommendations\cite{wang2024towards}.}

\begin{figure}[]
\centerline{\includegraphics[width=0.7\linewidth]{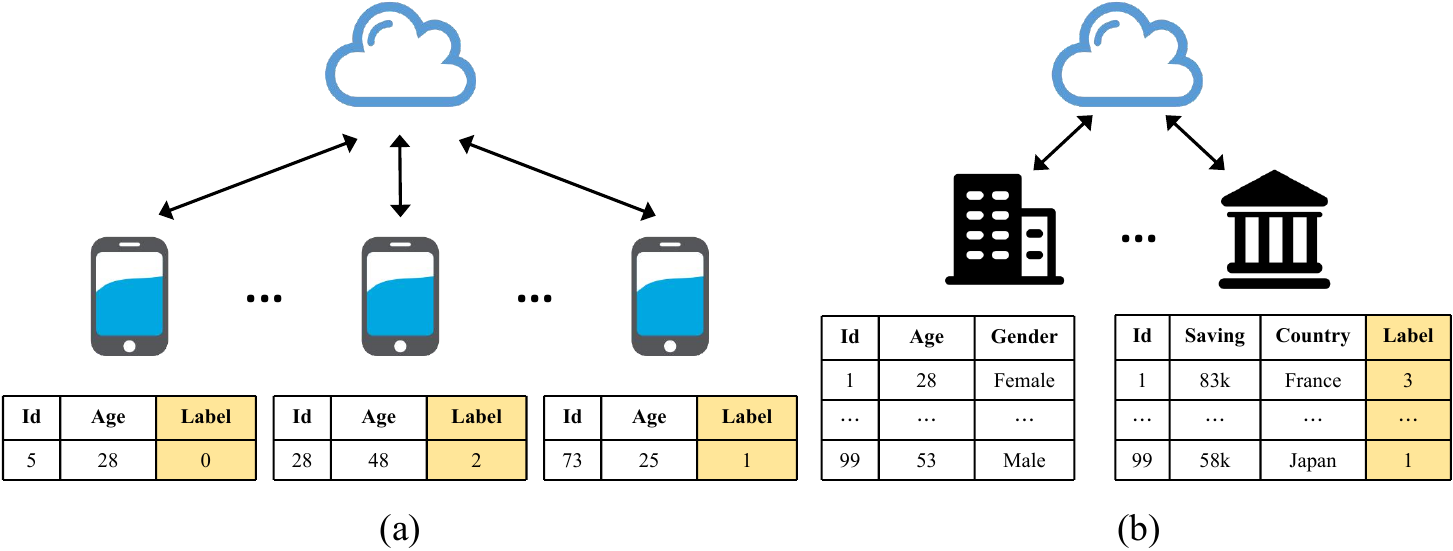}}
\caption{Data partitioning of HFL and VFL. (a) Horizontal partitioned data. (b) Vertical partitioned data.}
\label{fig:fl}
\end{figure}

Within the framework of federated learning, based on the distribution characteristics of local data, it is mainly divided into two scenarios: Horizontal Federated Learning (HFL) and Vertical Federated Learning (VFL), as shown in Fig. \ref{fig:fl}. In HFL, the local datasets of data owners have almost no intersection in the sample space, but there is a significant overlap in the feature space. In contrast, in VFL, local datasets have a large intersection in the sample space, but little overlap in the feature space.  
Each of these federated learning types has its applicable scenarios and advantages. VFL is particularly suitable for situations where different institutions hold different feature data of the same set of users; for example, in the financial sector, one institution may have the credit history of users, while another may have their transaction data. Through VFL, these institutions can collaborate to build more accurate risk assessment models without the need to directly exchange sensitive data. VFL has found wide applications in fields such as finance and healthcare.

Despite the advantages of VFL in protecting private data, recent studies have shown that it may still face risks of data privacy breaches \cite{wu2022federated}, especially through data reconstruction attacks. Such attacks reconstruct the original features of the training dataset by analyzing intermediate data during the VFL process, potentially leading to sensitive information leaks. 

\begin{figure}[!tbp]
\centerline{\includegraphics[width=0.5\linewidth]{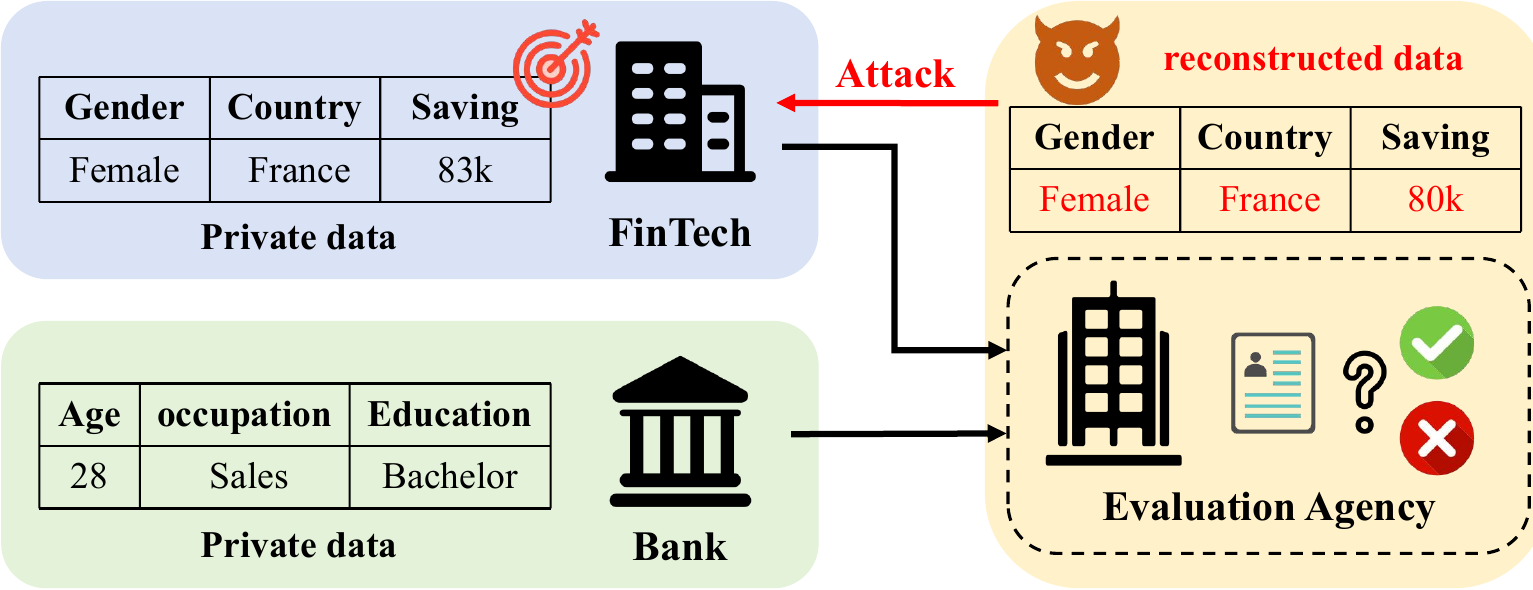}}
\caption{Illustration of a VFL data reconstruction attack, showing the bank and Fintech company with their bottom models and the evaluation agency with the top model and a bottom model. The agency conducts an attack on the FinTech company's model using VFL intermediate data to access private data while adhering to VFL protocols.}
\label{fig:scene}
\end{figure}

Fig. \ref{fig:scene} shows a VFL application scenario that is vulnerable to data reconstruction attacks \cite{chen2022graph}. In this scenario, a bank and a fintech company participate in a VFL for credit analysis, with each entity possessing a subset of user attributes. An assessment agency uses the labels of users to coordinate the training of the VFL model. However, the assessment agency wants to acquire the fintech company's private data and take it for its own use, so it launches a data reconstruction attack on the model owned by the fintech company. Without violating the VFL protocol, it uses the intermediate data from the VFL to reconstruct the private data owned by the bank, leading to the leakage of the fintech company's customer information and posing a significant security risk to the actual application of VFL.

Existing data reconstruction attacks on VFL, such as the generative regression network (GRN) method \cite{luo2021feature} and the gradient-based inversion attack (GIA) method \cite{jiang2022comprehensive}, largely draw from methods used in HFL (Horizontal Federated Learning), with a core focus on utilizing model information. However, these methods have certain limitations. 
\textcolor{black}{For example, the GRN method is primarily based on a white-box scenario, where the attacker reconstructs input data by accessing the passive party's model to calculate gradients in a multi-layer neural network. However, since passive parties in VFL typically do not share their models, this method is significantly constrained in practical applications. To address this limitation, the GIA method introduces a black-box attack scenario, where a proxy model is constructed, and data reconstruction is achieved by optimizing on this shadow model. Although the GIA method alleviates some limitations of the white-box scenario, it imposes strict requirements on model types, such as being applicable only to logistic regression (LR) models or neural networks without nonlinear activation functions in the output layer. Additionally, the GIA method requires individual optimization for each reconstruction, which further restricts its applicability and practicality due to its operational complexity and efficiency bottlenecks.}

It is crucial to go beyond the existing attack methods and explore a broader range of attack scenarios in order to gain a more thorough understanding of the potential threats to data privacy in VFL. 
The goal of this paper is to fully consider real-world attack scenarios in practical VFL environments and develop effective attack methods for different scenarios. To accomplish this, we have completely abandoned the traditional approach that relies on gradient information or model information and instead have opted to directly utilize the intermediate feature data generated in the VFL framework for attacks.
This method constructs an inverse net (InverNet) to effectively extract original data information from the intermediate features output from the target's model. Following this strategy, We have developed an innovative attack framework, called Unified InverNet Framework in VFL (UIFV), that is applicable to a variety of VFL scenarios with different adversary capabilities. UIFV overcomes the dependency on gradient information or model information of traditional attacks in VFL, thereby providing a more flexible and effective means for data reconstruction attacks in complex VFL environments.

Specifically, we make the following contributions in this paper.

\begin{itemize}
    \item \textbf{Exploration of VFL Data Reconstruction Risks}: We thoroughly investigate the risks of data reconstruction in VFL, analyze potential threats in different attack scenarios, and provide important insights for future defense measures.
    \item \textbf{Flexible Attack Framework}: We develop a framework UIFV that is applicable in various black-box attack scenarios for effective data reconstruction in VFL with different adversary's capabilities.
    \item \textbf{Stealth and Non-Intrusiveness}: The method and framework are designed to be stealthy and non-intrusive, allowing attacks without disrupting normal VFL operations and less likely to be detected.
    \item \textbf{High Attack Effectiveness}: We have conducted extensive experiments to evaluate the effectiveness of the proposed method and verified its higher attack accuracy compared to state-of-the-art methods.  
\end{itemize}

The remainder of this paper is organized as follows: Section \ref{sec:2} briefly reviews related works on Vertical Federated Learning (VFL) and associated attacks. Section \ref{sec:3} defines the problem. In Section \ref{sec:4}, we provide an overview of the UIFV attack framework. Section \ref{method} details the four scenarios within the UIFV attack framework. Section \ref{experiment} presents experimental results of UIFV in VFL, demonstrating the success of our attack. Finally, the paper is concluded with a summary in Section \ref{sec:7}.

\section{Related Work}\label{sec:2}

\subsection{Vertical Federated Learning}

The participants in a VFL framework consist of an active party and some passive parties \cite{liu2024vertical}. Each passive party owns a set of data features that are fed into a 
model (called a bottom model) for local training. The active party holds the label information and a top model in addition to its own feature set and bottom model. The active party coordinates the training process by concatenating the bottom model outputs as the input to the top model. This structure enables collaborative model training without the need to share original data, thus preserving the privacy of sensitive data on the participants.

As an important branch of federated learning, participants in VFL typically include one active party and several passive parties. Each passive party has a set of data features and trains a local bottom model. The active party, in addition to having its own feature set and bottom model, also holds label information and a top model. The active party coordinates the entire training process by passing the output of the bottom models as input to the top model. This structure enables collaborative model training without sharing raw data, thereby protecting the sensitive data privacy of participants.

VFL has extensive application prospects in fields such as finance, advertising, and healthcare. Kang et al.\cite{9826576} proposed a privacy-preserving VFL framework designed for financial applications, significantly improving the performance of credit loans. Li et al.\cite{li2022label} designed a label-protecting VFL framework that enhances advertising conversion rates. Fu et al.\cite{277244} made progress in the study of ductal carcinoma in situ (DCIS) using a VFL framework\cite{DBLP:conf/midp/Cruz-RoaBGGFGST14}. This paper provides a formal definition of VFL in section \ref{sec:3-1}.

\subsection{Attacks in VFL}

Although VFL has made significant progress in practical fields such as finance and advertising, its potential security vulnerabilities have raised widespread concerns, especially threats from within the VFL system, where one or more participants may attempt to attack others. The internal security issues of VFL can be divided into two main categories: one is attacks that disrupt the normal operation of VFL. For example, Liu et al.\cite{liucopur} used Projected Gradient Descent\cite{DBLP:conf/iclr/MadryMSTV18}(PGD) and feature flipping attacks to poison VFL predictions, and Chen et al.\cite{chenVFLbackdoor,CHEN2024103601} implanted backdoors during the VFL training phase, allowing passive parties to arbitrarily control prediction results during the inference phase. The other category involves attempts to obtain data from other participants, particularly private features or label information. Research on label attacks is relatively extensive. Li et al.\cite{li2022label} used direction and norm scoring methods to infer server labels based on distribution differences between positive and negative samples, and Zou et al.\cite{9833321} stole server labels through gradient inversion. In comparison, data reconstruction attacks pose a more serious threat in VFL, as attackers can reconstruct the original input data of other parties by analyzing intermediate gradients or model parameters, severely threatening data privacy. Existing studies, such as the Generative Regression Network (GRN) method\cite{luo2021feature} and the Gradient-based Inversion Attack (GIA) method\cite{jiang2022comprehensive}, have limited application in VFL scenarios, necessitating a more comprehensive analysis of the risks associated with VFL data reconstruction.

\subsection{Data Reconstruction Attack}
Current data reconstruction research can be categorized into gradient-based, model information-based, and feature-based methods.

\begin{table}[!tbp]
\centering
\adjustbox{max width=\textwidth}{
    % \begin{tabular}{p{3cm}p{3cm}p{1cm}p{1cm}p{1cm}p{1cm}p{1cm}p{1cm}c}
    \begin{tabular}{l@{\hspace{0.0pt}}c@{\hspace{1.5pt}}ccccccc}
    \toprule
        \multirow{3}{*}{\textbf{Method}} & \multirow{3}{*}{\textbf{Attack Type}} & \textbf{Requires } & \textbf{Requires } & \textbf{Needs Model} & \textbf{Modifies  } & \textbf{Requires} & \textbf{Training} & \multirow{2}{*}{\textbf{Training/Inference}} \\ 
    & & \textbf{Inference} & \textbf{Gradient} & \textbf{Structure and}  & \textbf{Model}& \textbf{Non-IID}& \textbf{data} & \\
    & & \textbf{Query}& \textbf{Query}& \textbf{Parameters}& \textbf{Architecture} & \textbf{data} & \textbf{Support$^*$} & \textbf{Support}\\  \midrule
    DGL\cite{zhu2019deep} & \multirow{4}{*}{Gradient-based} & - & \checkmark & \checkmark & - & - & - & Training \\   
    SQR\cite{wang2023reconstructing} &  & - & \checkmark & \checkmark & - & - & - & Training \\   
    CPA\cite{kariyappa2023cocktail} &  & - & \checkmark & \checkmark & - & - & - & Training \\       
    LOKI\cite{zhao2024loki} &  & - & \checkmark & \checkmark & \checkmark & - & - & Training \\       
    \midrule
    GRN\cite{luo2021feature} & \multirow{1}{*}{Model } & - & - & \checkmark & - & - & - & Training/Inference \\   
    GIA\cite{jiang2022comprehensive} &  information-based & - & - & $\circ$ & - & \checkmark & - & Training/Inference \\   \midrule
    Ginver\cite{yin2023ginver} & \multirow{2}{*}{Feature-based} & \checkmark & - & $\circ$ & - & - & - & Training/Inference \\   
    \textbf{UIFV} &  & $\circ$ & - & - & - & $\circ$ & \checkmark & Training/Inference \\ 
    \bottomrule
    \end{tabular}
}
\label{tab:Requirements}
\caption{\textcolor{black}{Comparison of Data Reconstruction Attack Methods and Their Requirements (\checkmark indicates the presence of a specific requirement or feature, - indicates that the requirement is not needed, and $\circ$ denotes optional or partially required conditions. $^*$ refers to whether the method supports only using a small amount of real leaked samples to assist the attack.)}}
\end{table}

\subsubsection{\textcolor{black}{Gradient-Based Methods}}

\textcolor{black}{Gradient-based methods achieve data reconstruction by leveraging the gradients generated during the training process of machine learning models. These methods require the attacker to access both the model and the gradients of the target data. Currently, these methods can be categorized into two main types:
The first type, represented by the pioneering work Deep Gradient Leakage (DGL) \cite{zhu2019deep}, focuses on initializing virtual data and calculating its gradients to achieve reconstruction. The attack process can be represented by the following equation:
\begin{equation} \label{eq:g} 
\hat{x}=\arg\min_x\|\nabla L(x,y,\theta)-\nabla W\|^2 
\end{equation}
Here, $\nabla L(x, y, \theta)$ represents the generated gradients, $\nabla W$ is the true gradients, $x$ is the virtual data, $\hat{x}$ is the reconstructed data, $L$ is the loss function, $y$ is the label, and $\theta$ is the model parameter. Recent studies, such as \cite{vero2022data} and \cite{wang2023reconstructing}, have further extended this approach with remarkable results. For instance, \cite{vero2022data} proposed TabLeak, a method specifically designed for tabular data, while \cite{wang2023reconstructing} used a tensor decomposition approach to reconstruct private samples through a single gradient query (In this paper, we refer to this method as SQR).}

\textcolor{black}{
The second type, represented by the LOKI approach in \cite{zhao2024loki}, primarily targets attacks on linear models. These methods directly compute gradients to recover the original data. The attack process can be described by the following equation:
\begin{equation}\label{eq:LLL} \hat{x} = \frac{\delta L}{\delta W^i} / \frac{\delta L}{\delta B^i} \end{equation}
Here, $\frac{\delta L}{\delta W^i}$ is the weight gradient, $\frac{\delta L}{\delta B^i}$ is the bias gradient of neuron $i$, and $x$ is the data that activates neuron $i$.}

\textcolor{black}{
However, gradient-based methods face a significant limitation: their performance degrades rapidly as the batch size increases, due to the mixing of gradients from different samples. To address this issue, \cite{kariyappa2023cocktail} employed Independent Component Analysis (ICA) to separate independent update signals and proposed the CPA method, successfully enabling reconstruction even with larger batch sizes. Additionally, \cite{zhao2024leak} systematically described how attackers could exploit gradient information in federated learning and provided detailed steps for implementing these attacks. 
It is worth noting that these gradient-based methods are primarily applicable to HFL architectures. In VFL, where access to gradient information is limited, these methods face substantial challenges.}

\subsubsection{\textcolor{black}{Model Information-Based Methods}}

\textcolor{black}{
Model-based reconstruction methods leverage the internal parameters of machine learning models to recover original training data. Unlike gradient-based methods, these approaches do not require access to gradient information during the training process but instead rely on access to the model and its internal parameters. The core idea is to initialize virtual data and compute its output on the model, optimizing it by comparing against the target output. The reconstruction process can be expressed as:
\begin{equation} 
\hat{x} = \arg\min_x \|f(x, \theta) - H\|^2 
\end{equation}
where $H$ is the true output of the target data on the model $f$. \cite{luo2021feature} proposed the GRN method to recover the passive party's data in VFL, which requires access to the passive party's model and its parameters. Building on GRN, \cite{jiang2022comprehensive} introduced the GIA method, which uses a small amount of known auxiliary data and their confidence scores to construct a shadow model that simulates the passive party's model, enabling data reconstruction without direct access.}

\textcolor{black}{
Essentially, these methods rely on the internal parameters of the model and adopt a white-box approach to optimize virtual data. However, due to concerns over data privacy and security, participants in VFL frameworks are often reluctant to share model details, making it highly challenging to access the passive party's model.
}

\subsubsection{\textcolor{black}{Feature-Based Methods}}

\textcolor{black}{
Feature-based methods leverage model outputs, such as Shapley values or intermediate features, to reconstruct data. These approaches do not require knowledge of model parameters or gradients. The core idea is to establish a relationship between the model's output features and its input, aiming to reconstruct the input data based on the output features. The attack process can be described as:
\begin{equation} 
\hat{x} = g(H, \hat{\theta}_g) \quad \text{where} \quad \hat{\theta}g = \arg \min{\theta_g} \left\| g(H, \theta_g) - x \right\|^2 \end{equation}
Here, $g$ is a function that maps $H$ back to $x$, and $\theta_g$ represents the parameters of the function $g$. In traditional machine learning, \cite{luo2022feature} was the first to reveal the risks of feature inference attacks in model explanations based on Shapley values, demonstrating that current explanation methods can lead to privacy leakage. }

\textcolor{black}{
In the field of split learning, some studies have also proposed attack methods targeting this issue. However, these methods have certain limitations. For example, the methods proposed by He et al. in \cite{he2019model,he2020attacking} heavily rely on auxiliary datasets to reconstruct user inputs in black-box settings. Additionally, the Ginver attack proposed in \cite{yin2023ginver} relies on Inference Queries, requiring one query for each gradient update of $g$. It is noteworthy that these methods have not yet been validated in VFL architectures. 
Our method, UIFV, is specifically designed to address the characteristics of VFL. It innovatively treats the passive party model in VFL as a black box and further reduces the prerequisites for launching attacks through two modules: Data Preparation and Model Preparation. This design significantly broadens the application scenarios of VFL data reconstruction attacks while enhancing their flexibility and adaptability.}

\section{Preliminaries}\label{sec:3}

\subsection{System Model}\label{sec:3-1}

Without loss of generality, we consider a VFL system with \( K \) participants, \( \mathbb{P}_1, \ldots, \mathbb{P}_K \), where \( K \geq 2 \). 
Each sample \( x_i = \{x_i^1, \ldots, x_i^{K}\} \) is a vector that comprises \( K \) sets of features each owned by one participant; i.e., \( x_i^k \) is the feature set provided by participant \( \mathbb{P}_k \).
The label set \( \{y_i\}_{i=1}^{N} \) can be viewed as a special feature that is typically owned by one of the participants, say \( \mathbb{P}_1 \), which is referred to as the active party, while the other participants are called passive parties. 

\begin{figure}[]
\centerline{\includegraphics[width=0.38\linewidth]{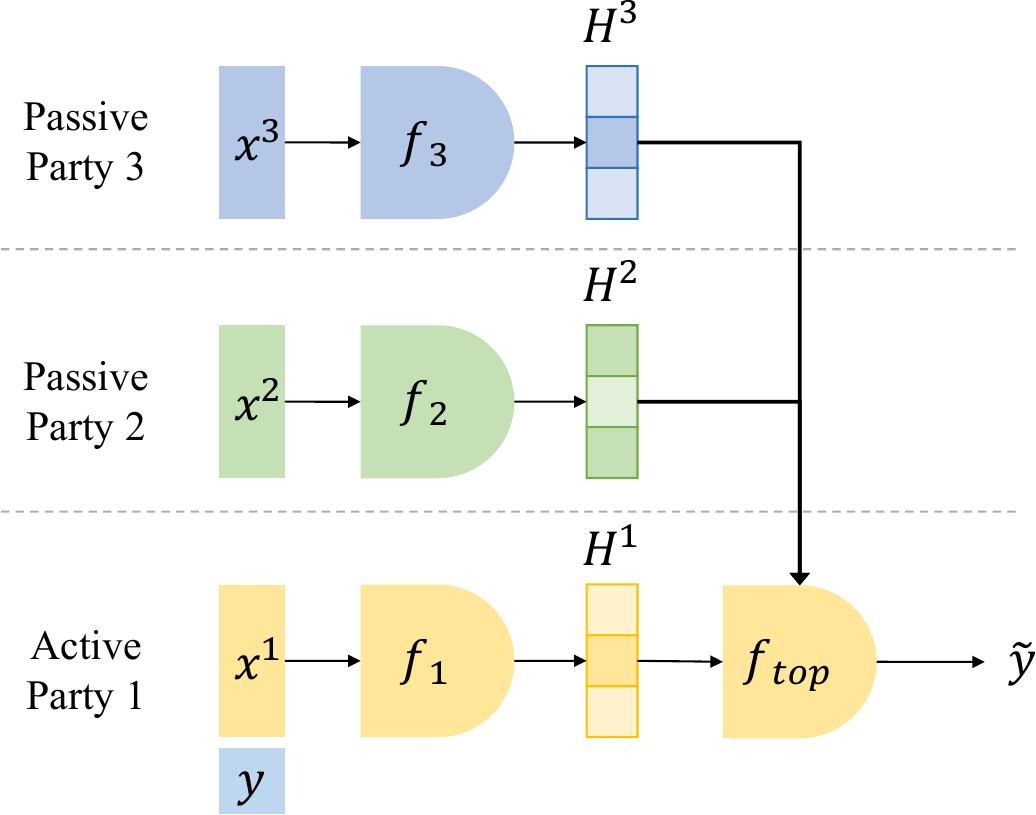}}
\caption{VFL Architecture Diagram with Three Participants}
\label{fig:systemModel}
\end{figure}

The VFL model can be represented as \( f_{\text{top}}(H^1, \cdots, H^{K}) \), where $H^k=f_{k}(\theta_k; x^k)$. 
$f_{top}( )$ is the top model controlled by the active party (owner of the labeled data), and $f_{k}(\theta_k; x^k)$ ($k=1, \cdots, K$) are the bottom models of the $K$ participants.  
For training the entire model, each participant $\mathbb{P}_k$ feeds the bottom model $f_{k}(\theta_k; x^k)$ with its own feature set $x^k_i, i=1, \cdots, N$ to generate the intermediate features $H^k$, which is then sent to the active party. The active party concatenates the intermediate features received from all participants to form the input to the top model and completes the forward propagation to generate an output, which is then used together with the label to calculate the loss function and determine the gradients. The top model is first updated based on the gradients, and then the partial gradients with respect to each bottom model are sent back to the participants to complete the backpropagation and update the bottom models. 
Therefore, the VFL model training can be formulated as 
\begin{equation}\label{eq1}
    \underset{\Theta}{ \min } \mathbb{E}_{(x, y) \sim \mathcal{D}} \mathcal{L}\left(f_{\text{top}}\left(H^1, \cdots, H^{K} \right), y\right),
\end{equation}
where \( \mathcal{D} \) is the training dataset, \( \mathcal{L} \) is the loss function, and $\Theta=\{\theta_1,\cdots,\theta_{K};\theta_{top}\}$ are VFL model parameters. A VFL framework with three participants (one active party and two passive parties) is illustrated in Fig. \ref{fig:systemModel}.

\subsection{Threat Model}

In this study, we assume that the adversary $\mathbb{P}_{\text{adv}}$ is the active party. One of the passive parties is the target of the attack, denoted as \( \mathbb{P}_{\text{target}} \). All participants strictly adhere to the VFL protocol.

\textbf{Adversary's objective.}
The goal of the attack is to acquire the private data \( x^{\text{target}} \) used by \( \mathbb{P}_{\text{target}} \) in VFL training. This objective is quite ambitious, fundamentally challenging privacy preservation in VFL. Naturally, this is also very difficult to achieve and nearly impossible in some practical scenarios. Therefore, the goal of the attack may be reduced to acquiring as much information about \( x^{\text{target}} \) as possible, for example, obtaining information about a specific column in \( x^{\text{target}} \).

\textbf{Adversary's capacity.} 
We assume that the attacker $\mathbb{P}_{\text{adv}}$ strictly follows the VFL protocol and cannot disrupt the normal operations on \( \mathbb{P}_{\text{target}} \); therefore, $\mathbb{P}_{\text{adv}}$ has no access to the bottom model at \( \mathbb{P}_{\text{target}} \). Since $\mathbb{P}_{\text{adv}}$ is on the active party, it can obtain the intermediate features $H^1, \cdots, H^K$ from all participants, including \( \mathbb{P}_{\text{target}} \). In different VFL scenarios, $\mathbb{P}_{\text{adv}}$ may have the following capabilities: possesses an auxiliary dataset with an identical and independent distribution (i.i.d.) as \( x^{\text{target}} \), make inference queries to the \( \mathbb{P}_{\text{target}} \)'s model, or obtain a minimal amount of private data samples used by \( \mathbb{P}_{\text{target}} \) for local training.  %The 

\begin{table}[!tbp]
    \centering
    \resizebox{0.45\textwidth}{!}{
    \begin{tabular}{p{4.5cm}ccc}
        \toprule
        \textbf{Setting} & \multicolumn{1}{m{0.8cm}}{\textbf{i.i.d. Data}} & \multicolumn{1}{m{1.8cm}}{\textbf{Inference Query}} &  \multicolumn{1}{m{2cm}}{\textbf{Minimal \( \mathbf{x^{\text{target}}} \)}}  \\ 
  \midrule

        Query Attack  & \checkmark  &  \checkmark & -  \\ 
        Data Passive Attack & \checkmark & - &  -  \\
        Isolated Query Attack & -  & \checkmark &  -  \\
        Stealth Attack & -   & - & \checkmark \\ 
    \bottomrule
    \end{tabular}
    }
    \caption{Adversary’s capability in our consideration (\checkmark: the adversary possess this capability; -: this capability is not necessary.)}
    \label{tab:advCap}
\end{table}

The adversary may launch a variety of attacks for data reconstruction based on the different capabilities that it has. Table \ref{tab:advCap} lists the attack scenarios and the required adversary capabilities.  
In the Query Attack scenario, the attacker possesses an i.i.d. auxiliary dataset and can make queries to $\mathbb{P}_{\text{target}}$. The Data Passive Attack only requires the attacker to own an i.i.d. dataset as the target's private data, while the Isolated Query Attack assumes the attacker can make queries to $\mathbb{P}_{\text{target}}$ but has no auxiliary dataset. In the Stealth Attack scenario, the attacker can neither make queries nor has i.i.d. data but has obtained a minimal amount of the target's private data used in training its bottom model.

\section{UIFV Framework Overview}\label{sec:4}

\textcolor{black}{We have developed a comprehensive framework, the Unified InverseNet Framework in VFL (UIFV). This framework enables attackers with varying capabilities, acting as the active party, to train an InverseNet and leverage the intermediate features of the passive party during normal VFL training or inference processes to reconstruct the private data of the passive participant.}

\begin{figure*}[!tbp]
\centerline{\includegraphics[width=0.66\linewidth]{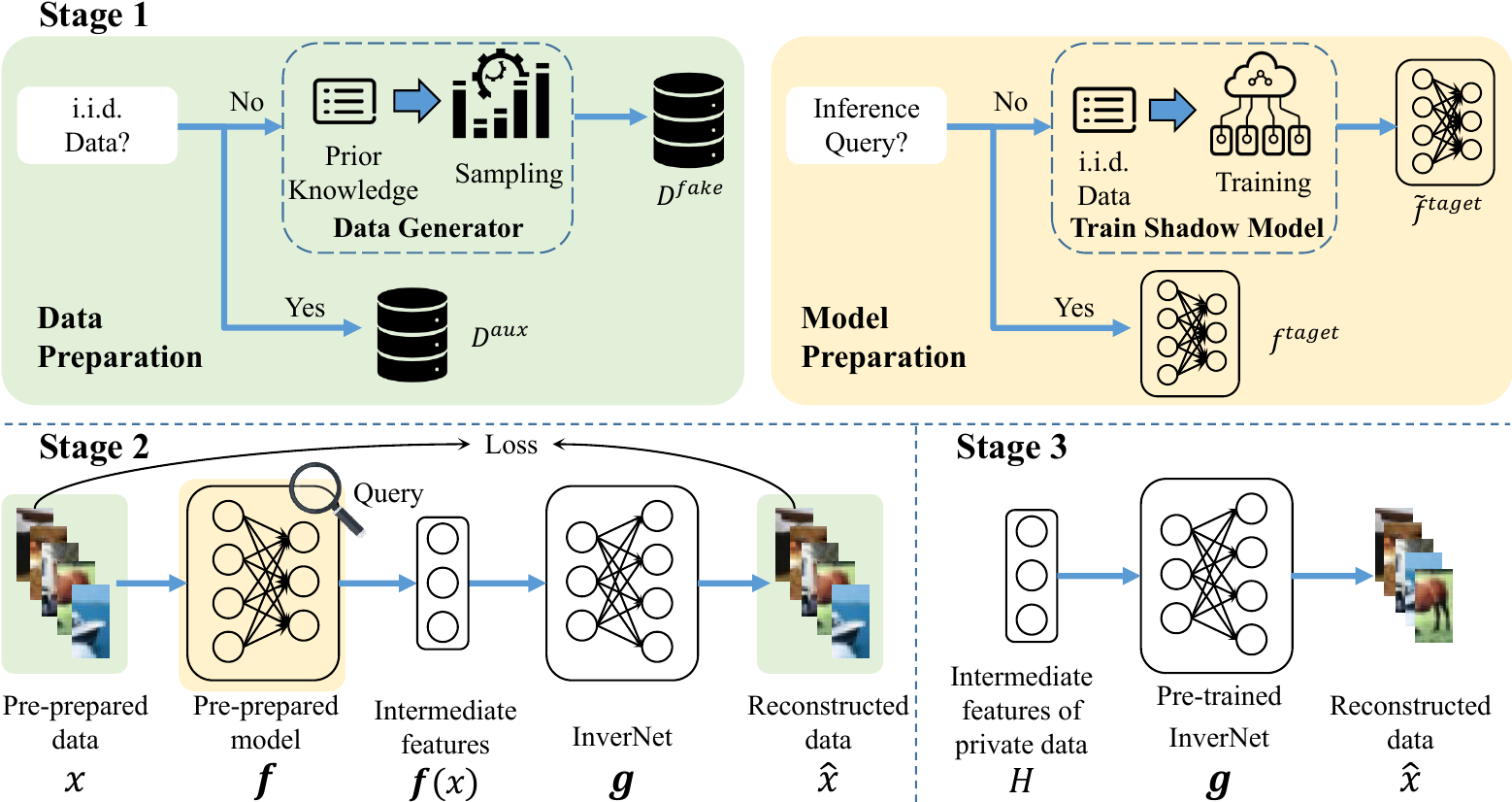}}
\caption{An overview of Unified InverNet Framework in VFL.}
\label{fig:overview-min}
\end{figure*}

\textcolor{black}{In the VFL framework, when we consider the $k$-th participant ($k \neq 1$) as our attack target, we treat the bottom model $f_k$ of participant $\mathbb{P}_k$ as a feature extractor. }The bottom model $f_k$ generates the intermediate features $H^k$ that are fed into the top model $f_{\text{top}}$. Considering that the intermediate layers of neural networks retain rich semantic information of input data, the proposed UIFV framework focuses on training an InverNet $g$ that establishes the relationship $g(f_k(x^{k})) = x^{k}$, which can then be used to upsample and reconstruct the private data $x^{k}$. The objective function for training the InverNet $g$ can be formulated as:
\begin{equation}\label{inverNet}
\textcolor{black}{\arg \min_{\theta_{g}} \left\| g(H^{k}) - x^{k} \right\|^2,}
\end{equation}
which indicates that the private data $x^k$ and the bottom model $f_k$ are required for training the InverNet. However, such information cannot be directly obtained in realistic VFL scenarios. Therefore, Data Preparation and Model Preparation are two key functional modules in the proposed UIFV framework that respectively prepare the model and data information needed for training the InverNet. 
\textcolor{black}{Based on whether these two modules are utilized, the attack scenarios are categorized into four types, which will be discussed in sections \ref{method}.}

The Data Preparation module may be implemented in different ways based on the attacker's capabilities. If the attacker possesses an i.i.d. auxiliary dataset $\tilde{x}^k$, it can be leveraged as a substitute for the private data $x^k$. If the attacker does not own such auxiliary data, the Data Preparation function can be realized through a data generator that utilizes the prior information of the original data (such as data distribution characteristics) to generate synthetic data $x^{\text{fake}}$.

Similarly, different methods can be employed by the Model Preparation module based on the attacker's capabilities. If the attacker can make inference queries to the target party, which is reasonable in the VFL architecture, then the model output $f_k(x^k)$ can be obtained from the queries. If the attacker cannot make query requests to the bottom model, then the attacker may train a shadow model $\tilde{f}_k$ to replace the model $f_k$ for training the InverNet. \textcolor{black}{Note that in the following sections, we will omit the index $k$ of the target party when there is no ambiguity.}

After completing the training of InverNet $g$ in UIFV, the intermediate features $H^{\text{target}}$ received from the target party can be passed to the InverNet $g$ to obtain an estimate $\hat{x}^{\text{target}}$ of the original private data $x^{\text{target}}$.

\begin{figure*}[tbp]
\centerline{\includegraphics[width=0.7\linewidth]{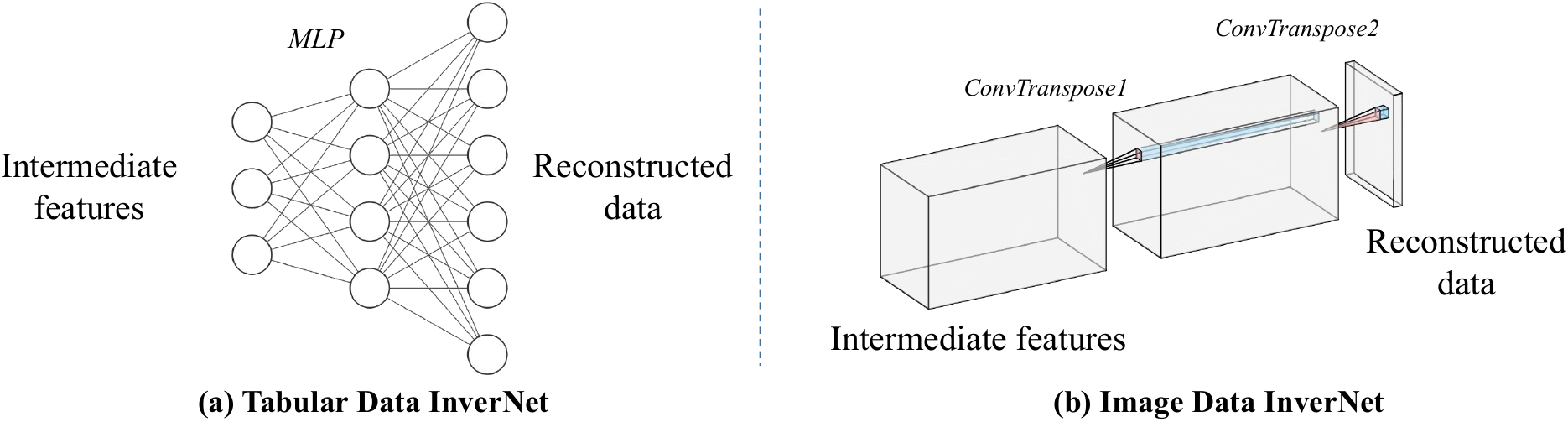}}
\caption{\textcolor{black}{InverNet Architecture for Different Data Types.}}
\label{fig:invernet}
\end{figure*}

As depicted in Fig. \ref{fig:overview-min}, the UIFV framework encompasses three distinct stages. The Data Preparation and Model Preparation functions are performed in the initial stage to prepare the pertinent data and model needed for training the InverNet, which is then conducted in the second stage. Then, in the final stage, the pre-trained InverNet is employed to reconstruct the private data used for model training by the target party. For different types of data, we have designed different InverNet architectures, as shown in Figure \ref{fig:invernet}.

\section{Four Attack Scenarios in UIFV} \label{method}

\subsection{Query Attack (QA)}\label{qa}
In the Query Attack scenario, the attacker has no access to the target party's model parameters and model gradient information during the training process, which is consistent with the black-box attack setting \cite{jiang2022comprehensive}. %,liu2021batch
On the other hand, the attacker can freely initiate query requests to the target party and possesses the auxiliary dataset \textcolor{black}{$D^{\text{aux}}$}
% $\tilde{x}^{\text{target}}$
, that is, i.i.d. with the target party's private data.

\begin{algorithm}[!t]
    \caption{Query Attack}
    \scriptsize % 缩小字体
    \label{alg:QA}
    \SetAlgoLined
    \SetKwFunction{QA}{QueryAttack}
    \SetKwFunction{trainNet}{TrainInverNet}
    \SetKwFunction{inv}{Inverse}
    \SetKwProg{Fn}{Function}{:}{}
    \Fn{\QA{$\textcolor{black}{\textcolor{black}{f_k}}$, $H^{\text{target}}$, $D^{\text{aux}}$}}{
        
        $g$=TrainInverNet($D^{\text{aux}}$ , $\textcolor{black}{\textcolor{black}{f_k}}$)   

        $\hat{x}_{\text{target}}$ = Inverse($g$, $H^{\text{target}}$)
        
        \KwRet $\hat{x}_{\text{target}}$
    }
    \BlankLine %空一行
    
    \Fn{\trainNet{$D^{\text{aux}}$, $\textcolor{black}{\textcolor{black}{f_k}}$}}{
   
        \BlankLine %空一行
    
        \While{$n<NIters$}{
        
            \textcolor{black}{Randomly sample $\tilde{x}_1,\tilde{x}_2...\tilde{x}_m$ from $D^{\text{aux}}$}

            % $H_i = f_{\text{target}}(x_i)$
            Obtain \( \tilde{x}_i \) by querying \(\textcolor{black}{\textcolor{black}{f_k}} \) with \( x_i \)
            
            $L(g)=\frac{1}{m} \sum_{i=1}^m \left\| g(H_i) - x_i \right\|^2$
            
            $\theta_g^{(n+1)}=\theta_g^{(n)}-\epsilon \frac{\partial L(g^{(n)})}{\partial \theta_g^{(n)}}$
        
            $n+=1$
        }
        \KwRet $g^{(\text{NIters})}$
    }

    \BlankLine %空一行
    
    \Fn{\inv{$g$, $H^{\text{target}}$}}{
        $\hat{x}_{\text{target}}$ = $g(H^{\text{target}})$
        
        \KwRet $\hat{x}_{\text{target}}$
    }

\end{algorithm}

\begin{equation}\label{QAloss}
\textcolor{black}{g = \arg \min_{\theta_{g}} \frac{1}{m} \sum_{i=1}^m \left\| g(\tilde{H}_i) - \tilde{x}_i \right\|^2}
\end{equation}

In this attack scenario, since the attacker owns an auxiliary dataset and is able to query the target model, the UIFV framework requires no work for data and model preparation in the first stage. In the second stage, auxiliary data $\tilde{x}$ is used to initiate query requests to the target party to obtain the intermediate features $\tilde{H}$ output by the target model $f_{k}$. Subsequently, the obtained $\tilde{H}$ and the corresponding $\tilde{x}$ are used to train InverNet $g$, as shown in \ref{QAloss}, where $m$ represents the number of samples in the dataset. Then, in the final stage of UIFV, the trained InverNet $g$ can be used for the reconstruction of the target data $x^{target}$. 
The complete attack algorithm is detailed in Algorithm \ref{alg:QA}.

\subsection{Data Passive Attack (DPA)}

In the scenario of Data Passive Attack, the attacker's capabilities are limited to only possessing the i.i.d. auxiliary dataset $\mathcal{D}^{\text{aux}}$ without the ability to query the target party. This could be because the target party only participates in the VFL training process but not the inference stage.
In such an attack scenario, Under this assumption, the Model Preparation function in the UIFV framework needs to construct a shadow model $\tilde{f}_k$ that mimics the behavior of the target model.

One approach to building a shadow model is to recover the structure and parameters of the target model by querying the black-box model, allowing the shadow model to mimic the behavior of the target model \cite{oh2019towards,tramer2016stealing,wang2018stealing}. However, since we cannot query the target model, we turn to the VFL architecture and attempt to build a model that behaves similarly to the target model within the VFL framework. In this process, 
we require the cooperation of other participants, querying them with i.i.d. data to obtain the corresponding intermediate features.
The optimization objective of the shadow model is
\begin{equation}\label{DPloss}
\textcolor{black}{\arg \min_{{\tilde{f}_k}} \frac{1}{m} \sum_{i=1}^m \mathcal{L}\left(f_{\text{top}}\left(H^1,\cdots,\tilde{f}_k(x_i),\cdots,H^K \right), y_i\right),  }
\end{equation}
where $m$ represents the number of samples in the dataset, and $H^i (i \neq k)$ is the intermediate features output of other participants, which remains constant during the training process. $\mathcal{L}$ is consistent with the loss function of the top model. This means that the training of $\tilde{f}_k$ is guided by the supervision of $f_{\text{top}}$.

\begin{algorithm}[!t]
    \caption{Data Passive Attack}
    \scriptsize % 缩小字体
    \label{alg:DP}
    \SetAlgoLined
    \SetKwFunction{DP}{DataPassiveAttack}
    \SetKwFunction{SD}{TrainShadowModel}
    \SetKwFunction{inv}{Inverse}
    \SetKwProg{Fn}{Function}{:}{}
    \Fn{\DP{$\textcolor{black}{f_k}$, $H^{\text{target}}$, $D^{\text{aux}}$}}{

        $\tilde{f}_k$ = TrainShadowModel($D^{\text{aux}}$, $f_{\text{top}}$)
    
        % \BlankLine %空一行
        
        $g$ = TrainInverNet($D^{\text{aux}}$ , $\tilde{f}_k$)   

        $\hat{x}_{\text{target}}$ = Inverse($g$, $H^{\text{target}}$)
        
        \KwRet $\hat{x}_{\text{target}}$
    }
    \BlankLine %空一行
    
    \Fn{\SD{$D^{\text{aux}}$ , $\textcolor{black}{f_k}$}}{

        \While{$n<NIters$}{
        
            % Randomly sample $x_1,x_2...x_k$ from $x$
            \textcolor{black}{Randomly sample $\tilde{x}_1$, $\tilde{x}_2$, ... ,$\tilde{x}_m$ and labels $\tilde{y}_1$, $\tilde{y}_2$, ... ,$\tilde{y}_m$ from $D^{\text{aux}}$}

            \BlankLine %空一行
            
            $\hat{y}_i$ = $f_{\text{top}}\left(H^1,\cdots,\tilde{f}_k(\tilde{x}_i),\cdots,H^K \right)$

            \BlankLine %空一行
            
            \textcolor{black}{$L(\tilde{f}_k)=\tfrac{1}{m} \sum_{i=1}^m \tilde{y}_i \hat{y}_i + (1-\tilde{y}_i)(1-\hat{y}_i)$}
    
            \BlankLine %空一行
            
            % $\tilde{f}_{\text{target}}^{(n+1)}=\tilde{f}_{\text{target}}^{(n)}-\epsilon \frac{\partial L(\tilde{f}_{\text{target}}^{(n)})}{\partial \tilde{f}_{\text{target}}^{(n)}}$

            $\theta_{\tilde{f}_k}^{(n+1)}=\theta_{\tilde{f}_k}^{(n)}-\epsilon \frac{\partial L(\tilde{f}_k^{(n)})}{\partial \theta_{\tilde{f}_k}^{(n)}}$
        
            $n+=1$
        }
        \KwRet $\tilde{f}_k^{({\text{NIters})}}$
    }
\end{algorithm}

Once the shadow model $\tilde{f}_k$ is trained in the first stage, it can be used in the role of $\textcolor{black}{f_k}$ together with the dataset $\mathcal{D}^{\text{aux}}$ to train the InverNet $g$ in the second stage and then to reconstruct the private data in the third stage of the UIFV framework, as in the Query Attack scenario. The complete attack algorithm in the Data Passive Attack scenario is shown in Algorithm \ref{alg:DP}.

\subsection{Isolated Query Attack (IQA)}
\begin{algorithm}[!t]
    \caption{Isolated Query Attack}
    \scriptsize % 缩小字体
    \label{alg:IQA}
    \SetAlgoLined
    \SetKwFunction{IQA}{IsolatedQueryAttack}
    \SetKwFunction{DG}{DataGeneration}
    \SetKwFunction{inv}{Inverse}
    \SetKwProg{Fn}{Function}{:}{}
    \Fn{\IQA{$\textcolor{black}{f_k}$, $H^{\text{target}}$, $T$}}{

        $D^{\text{fake}}$ = DataGeneration($T$)
    
        % \BlankLine %空一行
        $g$ = TrainInverNet($D^{\text{fake}}$ , $\textcolor{black}{f_k}$)   

        $\hat{x}_{\text{target}}$ = Inverse($g$, $H^{\text{target}}$)
        
        \KwRet $\hat{x}_{\text{target}}$
    }   
    
\end{algorithm}

In the Isolated Query Attack scenario, we assume that the attacker can freely initiate queries to the target party but does not have the i.i.d. data. Facing this situation, in the first stage of the UIFV framework, the Data Preparation module implements a Data Generator to create a set of fake data that are then used for querying the target part to obtain the intermediate features.

The most straightforward approach to creating fake data is random generation; for example, sampling pure noise from a standard Gaussian distribution as in \cite{he2019model} for image data. However, tabular data, which is significantly different in distribution from image data, consists of heterogeneous features and lacks spatial or semantic relationships, making it more complex to discover and utilize relationships \cite{borisov2022deep}. Therefore, using randomly generated data for query requests may lead to poor reconstruction results, as verified in Section \ref{sec:DG}.

To address this issue, we introduce prior knowledge to guide the process of generating random data, aiming to enhance the quality of data reconstruction. For image data, we focus on generating smoother random samples by reducing the influence of noise and outliers in random data through sampling from a standard Gaussian distribution. For tabular data, based on the analysis in \cite{borisov2022deep}, we utilize the header information of the target dataset to construct pseudo data, making the fabricated data closer to the distribution of real data. For categorical variables, we adopt a one-hot encoding approach, randomly selecting a category to set its corresponding column to 1, while keeping other columns at 0. For continuous variables, we first estimate the range of values based on experience and then perform random sampling within the estimated range using a uniform distribution. We call this process the data generation module.

The generated fake data $\mathcal{D}^{\text{fake}}$ will be used in the second stage to train InverNet $g$. Then, in the final stage of the UIFV framework, the trained InverNet $g$ will be used to reconstruct private data. The complete attack algorithm in the Isolated Query Attack scenario is shown in Algorithm \ref{alg:IQA}.

\subsection{Stealth Attack (SA)}
\begin{algorithm}[!t]
    \caption{Stealth Attack}
  \scriptsize % 缩小字体
    % \setstretch{0.9} % 调整行间距
    \label{alg:SA}
    \SetAlgoLined
    \SetKwFunction{SA}{StealthAttack}
    \SetKwFunction{trainNetfrom}{\textcolor{black}{TrainInverNetWithLeakedData}}
    \SetKwFunction{inv}{Inverse}
    \SetKwProg{Fn}{Function}{:}{}
    \Fn{\SA{$H^{\text{target}}$, $D^{\text{leak}}$}}{
        
        \textcolor{black}{$g$=TrainInverNetWithLeakedData($D^{\text{leak}}$)   }

        $\hat{x}_{\text{target}}$ = Inverse($g$, $H^{\text{target}}$)
        
        \KwRet $\hat{x}_{\text{target}}$
    }
    \BlankLine %空一行
    
    \Fn{\trainNetfrom{$D^{\text{leak}}$}}{

        $g^{(0)} = $ Init()

        \BlankLine %空一行
    
        \While{$n<NIters$}{
        
            \textcolor{black}{Obtain leaked samples $x_1^{\text{target}},x_2^{\text{target}}...x_m^{\text{target}}$ from $D^{\text{leak}}$}
            
            \textcolor{black}{Prepare $H_i^{\text{target}}$ data corresponding to $x_i^{\text{target}}$
            % $H_i = f_{\text{target}}(x_i)$
            , where $m \ll |H^{\text{target}}|$ }
            % with $|H^{\text{target}}| $ denoting the size of $H^{\text{target}}$.
            
            $L(g)=\frac{1}{m} \sum_{i=1}^m \left\| g(H_i^{\text{target}}) - x_i^{\text{target}} \right\|^2$
            
            $\theta_g^{(n+1)}=\theta_g^{(n)}-\epsilon \frac{\partial L(g^{(n)})}{\partial \theta_g^{(n)}}$
        
            $n+=1$
        }
        \KwRet $g^{(\text{NIters})}$
    }

\end{algorithm}

In the previous three attack scenarios, the attacker can possess an i.i.d. auxiliary dataset and/or make queries to obtain the intermediate features but does not know about the target party's private information. In this scenario, even if the attacker loses both the auxiliary data and the query capabilities, data reconstruction is still possible if the attacker \( \mathbb{P}_{\text{adv}} \) may acquire a minimal amount of the target party's private data \textcolor{black}{$D^{\text{leak}}(D^{\text{leak}}\subset x^{\text{target}})$} and explicitly knows that these data have already been used in the VFL process. For instance, \( \mathbb{P}_{\text{adv}} \) might collude with some internal employees of the target party (whose data are jointly maintained by both parties) to secretly acquire a small subset of data, which is considered possible as has been noted in some studies on analogous VFL scenarios, such as \cite{zeng2023narcissus}. The knowledge of some private data used by the target party allows the attacker to start from the second stage of the UIFV framework to directly train InverNet $g$ with the objective function simplified to \ref{inverNet-2}: 
\begin{equation}\label{inverNet-2}
\textcolor{black}{\arg \min_{\theta_{g}} \left\| g(H^{\text{target}}_{\text{sub}}) - x^{\text{target}}_{\text{sub}} \right\|^2}
\end{equation}
\textcolor{black}{In this equation, $x^{\text{target}}_{\text{sub}}$ signifies the secretly acquired data, while $H^{\text{target}}_{\text{sub}}$ represents the corresponding intermediate features of these secret data.} Then, the trained $g$ will be used in the third stage to complete private data reconstruction. The complete attack algorithm in the Stealth Attack scenario is shown in Algorithm \ref{alg:SA}.

\section{Experiments} \label{experiment}

\subsection{Experiment Setting}

In the following content, we will describe our experimental design, focusing on two parts: the datasets used and the models implemented.

\subsubsection{Datasets}

In our experiments, we employed four public datasets: Bank marketing analysis \cite{moro2014data}\footnote{\url{https://archive.ics.uci.edu/dataset/222/bank+marketing}} (Bank), Adult income \cite{misc_adult_2}\footnote{\url{https://archive.ics.uci.edu/dataset/2/adult}} (Income), Default of credit card clients \cite{yeh2009comparisons}\footnote{\url{https://archive.ics.uci.edu/dataset/350/default+of+credit+card+clients}} (Credit) and CIFAR10 \cite{krizhevsky2009learning}\footnote{\url{https://www.cs.toronto.edu/~kriz/cifar.html}}, to evaluate our methods. These datasets range from banking marketing analysis to image recognition, each with its unique features and challenges. For data preparation, continuous columns in tabular datasets were scaled, and discrete columns were one-hot encoded. We have summarized the evaluated datasets in Table \ref{tab:Dataset}. 

\begin{table}[tb]
\centering
\resizebox{0.45\textwidth}{!}{
\begin{tabular}{l@{\hspace{2pt}}c@{\hspace{6pt}}c@{\hspace{6pt}}c@{\hspace{6pt}}c}
\toprule
\textbf{Dataset} & \textbf{Bank} & \textbf{Income} & \textbf{Credit} & \textbf{CIFAR10}\\
\midrule
Sample Num. & $41,188$ & $32,561 $& $30,000$ & $60,000$ \\
Feature Num. & $20$ & $14$ & $23$ & $32\times32\times3$ \\
Class Num. & $2$ & $2$ & $2$ & $10$\\
Accuracy on VFL &$ 0.9153$ & $0.8417$ & $0.8322$ & $0.7493$ \\
AUC on VFL & $0.8254$ & $0.8969$ & $0.7844$ & - \\
\bottomrule
\end{tabular}
}
\caption{Dataset used in our experiments.}
\label{tab:Dataset}
\end{table}

\textcolor{black}{The Bank dataset, derived from a Portuguese banking institution's direct marketing campaigns, encompasses 41188 samples with 20 features, including 10 discrete features, aimed at determining the likelihood of clients subscribing to a term deposit. The features include age, job, marital status, and education. 
The Income dataset, also known as the "Census Income" dataset, includes 32561 instances with 14 features, including 8 discrete features such as work class, and education, and it aims to predict if an individual's income surpasses \$50,000 per annum using census data. 
The Credit dataset, sourced from Taiwan, comprises 30,000 instances with 23 features, including 9 discrete features such as credit amount, gender, education, and marital status, and aims to predict the probability of default payments in credit card clients using various data mining methods.}
The CIFAR10 dataset is a widely-used public dataset for computer vision research, containing 60,000 color images with a resolution of 32x32 pixels, divided into ten categories with 6,000 images in each category. 
Additionally, before training the models, we scaled the continuous columns in the tabular datasets to the range of [-1, 1], while the discrete columns were encoded using one-hot encoding. For image data, no preprocessing was performed. 
\textcolor{black}{
We used 80\% of the tabular data to train the VFL model, with the remaining 20\% serving as the target for data reconstruction attacks. For the CIFAR10 dataset, we used the training set consisting of 50,000 images to train the VFL model, and the test set with 10,000 images was used as the target for data reconstruction attacks.}

\subsubsection{Models}

For simplicity, we chose to focus on a two-party VFL setup in our experiment. Theoretically, this framework can be expanded to scenarios with any number of participants. In our attack scenario, VFL involves two roles: one is the active party, which plays the role of the adversary and possesses a complete top model and a bottom model; the other is the passive participant, serving as the target of the attack, equipped only with a bottom model. Regarding data splitting, unless specifically stated otherwise, it is generally assumed that the active and passive parties equally share the data. For tabular data, discrete and continuous data each constitute half; for image data, each image is bisected along the central line, with both parties holding half, but only the active party possesses the data labels. In terms of model construction, for processing tabular data, both parties use a three-layer fully connected neural network as the bottom model. The top-layer model is also composed of a three-layer fully connected network, with each layer incorporating a ReLU activation function. For models processing image data, both parties employ a network comprising two convolutional layers and one pooling layer as the bottom model, while the top model consists of four convolutional layers and two fully connected layers, with each layer also integrating a ReLU activation function. Upon applying this VFL architecture to four different datasets, we achieved the training performance results as shown in Table \ref{tab:Dataset}.

\textcolor{black}{
The InverNet for all bottom models is consistent with the architecture of the respective bottom model. 
For tabular data, the InverNet uses a three-layer fully connected neural network; for image data, the model employs two transposed convolutional layers, with a ReLU activation function between each layer. Model details are provided in table \ref{tab:model_architecture}.}

\begin{table}[ht]
\centering
\adjustbox{max width=\textwidth}{
\begin{tabular}{ccccc}
\toprule
 & \textbf{Bank} & \textbf{Income} & \textbf{Credit} & \textbf{CIFAR10} \\
\hline
\textbf{Bottom} & MLP & MLP & MLP & Conv2d(3$\to$32,kernel=3,padding=1) \\
 \textbf{Model} & (input\_dim,300,100,100) & (input\_dim,300,100,100) & (input\_dim,300,100,100) & Conv2d(64$\to$64, kernel=3,padding=1) \\
 &  &  &  & MaxPool2d(kernel=2,stride=2) \\
\midrule
\multirow{2}{*}{\textbf{InverNet}} & MLP& MLP & MLP & ConvTranspose2d(64$\to$64,kernel=3,padding=1) \\
 & (100,100,300,input\_dim)  & (100,100,300,input\_dim)  & (100,100,300,input\_dim)  & ConvTranspose2d(64$\to$3,kernel=3,padding=1) \\
\bottomrule
\end{tabular}
}
\caption{\textcolor{black}{Model architectures for different datasets.}}
\label{tab:model_architecture}
\end{table}

\subsection{Evaluation Metrics}\label{eva}
In our work, we evaluated two categories of data: tabular and image data, employing distinct metrics for each category.
% In our work, we assess two types of data: tabular and image, using different metrics for each.

For image data, we adopt two widely recognized metrics: Peak Signal-to-Noise Ratio (PSNR) and Structural Similarity Index (SSIM) \cite{wang2004image}. 
PSNR quantifies image errors by calculating the mean squared error between origin and attack images, with higher values indicating lower quality degradation. SSIM evaluates image quality based on structural information, brightness, and contrast, ranging from 0 to 1, with 1 indicating perfect similarity.

In evaluating tabular data, previous studies \cite{luo2021feature,luo2022feature} have used training loss or distance measures to assess reconstruction accuracy. However, these methods may not align with real attack scenarios, which focus on whether reconstructed categories match the actual ones. To address these issues, we adopted the metrics proposed in \cite{vero2022data}. Considering the characteristics of tabular data, we separate the treatment of categorical and continuous features. For vector $x$ and its reconstruction vector $\hat{x}$, the accuracy metric is defined as follows:
\begin{align}
\text{accuracy}(x, \hat{x}) :=  \frac{1}{M + L} \bigg( \sum_{i=1}^{M} \mathbb{I}\{x^{D}_{i} = \hat{x}^{D}_{i}\} \notag 
 + \sum_{i=1}^{L} \mathbb{I}\{\hat{x}^{C}_{i} \in [x^{C}_{i} - \varepsilon, x^{C}_{i} + \varepsilon]\} \bigg),
\end{align}
where $M$ and $L$ denote the number of discrete variables and continuous variables in vector $x$. The indicator function \(\mathbb{I}\) checks for equality in categorical features and for the continuous features being within an epsilon range \(\varepsilon\).

\subsection{Performance Evaluation and Comparison}

\textcolor{black}{In our VFL data reconstruction attack experiments, we conducted a comprehensive comparison between the proposed UIFV method and the latest state-of-the-art methods, using datasets including Bank, Income, and Credit. We evaluated the UIFV method in four different scenarios and ensured that the compared methods utilized the same architecture and consistent experimental settings as UIFV. For the GIA method \cite{jiang2022comprehensive} and the Ginver method \cite{yin2023ginver}, we adopted black-box attack versions where the attacker does not know the specific structure and parameters of the model. Additionally, we included a random guessing baseline method to evaluate the inherent performance of random predictions. During the evaluation, we applied the metrics defined in Section \ref{eva}, setting the $\varepsilon$ value for continuous features to 0.2 and the batch size to 64. Detailed results can be found in Table \ref{tab:eva}.}

\begin{table}[htbp]
\centering
\adjustbox{max width=0.38\textwidth}{
    \begin{tabular}{lccc}
    \toprule
    \textbf{Method} & \textbf{Bank} & \textbf{Income} & \textbf{Credit} \\
     \midrule
    DGL\cite{zhu2019deep} & $13.63 \pm 1.00$ & $33.20 \pm 0.74$ & $16.88 \pm 1.80$ \\
    SQR\cite{wang2023reconstructing} & $27.53 \pm 0.36$ & $22.05 \pm 0.50$ & $15.50 \pm 0.82$ \\
    CPA\cite{kariyappa2023cocktail} & $29.00 \pm 2.83$ & $33.33 \pm 2.49$ & $17.27 \pm 2.26$ \\
    LOKI$^*$\cite{zhao2024loki} & $14.72 \pm 0.74$ & $0.34 \pm 0.22$ & $0.00 \pm 0.00$ \\    
    % VFLRecon\cite{zhu2024vulnerabilities} & $98.67 \pm 0.05$ & $97.69 \pm 0.08$ & $98.37 \pm 0.05$ \\
    GRN\cite{luo2021feature} & $30.20 \pm 6.68$ & $41.12 \pm 8.52$ & $53.82 \pm 25.84$ \\
    GIA\cite{jiang2022comprehensive} & $55.93 \pm 1.81$ & $18.87 \pm 8.27$ & $41.07 \pm 2.78$ \\
    % 78.23	2.23	80.91	2.07	69.44	1.92
    Ginver\cite{yin2023ginver} & $78.23 \pm 2.23$ & $80.91 \pm 2.07$ & $69.44 \pm 1.92$ \\
    Random & $21.02 \pm 0.04$ & $11.14 \pm 0.07$ & $13.88 \pm 0.06$ \\
    \hline
    UIFV-QA & $97.96 \pm 0.11$ & $98.49 \pm 0.04$ & $98.19 \pm 0.13$ \\
    UIFV-DPA & $95.74 \pm 0.38$ & $80.80 \pm 0.91$ & $96.00 \pm 0.46$ \\
    UIFV-IQA & $66.17 \pm 1.43$ & $94.81 \pm 0.54$ & $44.25 \pm 1.61$ \\
    UIFV-SA & $90.07 \pm 0.11$ & $72.79 \pm 3.40$ & $93.83 \pm 0.20$ \\
    \bottomrule
    \end{tabular}
}
\caption{\textcolor{black}{Performance comparison with state-of-the-art methods on the Bank, Income, and Credit datasets. ($^*$Note: Unlike other methods, LOKI achieves 100$\%$ input recovery upon success, with accuracy defined as the proportion of successfully reconstructed data in the datasets.)}}\label{tab:eva}
\end{table}

\textcolor{black}{Due to differences in attack assumptions among the methods (as detailed in Table \ref{tab:Requirements}), it is challenging to directly compare our method with others. However, overall, the UIFV method achieved relatively high attack success rates under the weakest attack assumptions. 
When conducting a lateral comparison within specific scenarios, only GIA and Ginver overlap with UIFV in terms of attack scenarios: the experimental setting of GIA aligns with the UIFV-DPA scenario, but its performance across the three datasets is significantly lower than that of UIFV. 
Ginver's experimental scenario was more similar to UIFV-IQA, and their performances were comparable, with each method excelling in different aspects. However, Ginver was less flexible and applicable than UIFV.}

\textcolor{black}{We also observed that gradient-based attack methods, such as DGL and SQR, performed poorly across the three datasets, primarily due to the batch size being set to 64, which significantly impacted their attack performance. Similarly, model information-based attack methods, such as GRN and GIA, also exhibited poor performance on tabular data. This is because tabular data typically features high dimensionality and low correlation, making the optimization problem non-convex and complex, which often leads to local optima and reduces the likelihood of successful optimization.}

\subsubsection{\textcolor{black}{Performance on Tabular Datasets}}

\begin{table*}[!tbp]
\centering
\resizebox{0.6\textwidth}{!}{

\begin{tabular}{lcccccc}
\toprule
\textbf{Dataset} & \textbf{Evaluation} & \textbf{UIFV-QA} & \textbf{UIFV-DPA} & \textbf{UIFV-IQA} & \textbf{UIFV-SA} \\ 
\midrule
\multirow{3}{*}{Bank} & Accuracy & $98.0 \pm 0.1$ & $95.7 \pm 0.4$ & $66.2 \pm 1.4$ & $90.1 \pm 0.1$ \\
 & Discrete Acc & $99.5 \pm 0.0$ & $99.0 \pm 0.1$ & $87.8 \pm 2.1$ & $94.9 \pm 0.3$ \\
 & Continuous Acc & $96.4 \pm 0.2$ & $92.4 \pm 0.8$ & $44.6 \pm 1.2$ & $85.2 \pm 0.3$ \\
\midrule
\multirow{3}{*}{Income} & Accuracy & $98.5 \pm 0.0$ & $80.8 \pm 0.9$ & $94.8 \pm 0.5$ & $72.8 \pm 3.4$ \\
 & Discrete Acc & $98.4 \pm 0.0$ & $81.1 \pm 0.9$ & $98.1 \pm 0.6$ & $71.4 \pm 2.4$ \\
 & Continuous Acc & $98.8 \pm 0.1$ & $79.8 \pm 1.2$ & $81.7 \pm 0.7$ & $78.3 \pm 8.6$ \\
\midrule
\multirow{3}{*}{Credit} & Accuracy & $98.2 \pm 0.1$ & $96.0 \pm 0.5$ & $44.3 \pm 1.6$ & $93.8 \pm 0.2$ \\
 & Discrete Acc & $98.5 \pm 0.1$ & $97.1 \pm 0.5$ & $75.2 \pm 4.8$ & $92.2 \pm 0.3$ \\
 & Continuous Acc & $98.0 \pm 0.2$ & $95.3 \pm 0.6$ & $26.6 \pm 1.8$ & $94.8 \pm 0.2$ \\
\bottomrule
\end{tabular}
}
\caption{\textcolor{black}{Reconstruction Performance of UIFV on Discrete and Continuous Features Across Four Scenarios.}}
\label{tab:evaDisAndCon}
\end{table*}

\textcolor{black}{To further analyze the performance of UIFV in tabular data reconstruction, we divided the data from three datasets into two categories: discrete data and continuous data. As shown in Table \ref{tab:evaDisAndCon}, UIFV achieves significantly higher reconstruction accuracy on discrete data (encoded with one-hot) compared to continuous data across all four scenarios. For discrete data, UIFV consistently achieves over 70\% reconstruction accuracy in all scenarios. For continuous data, except in the IQA scenario, UIFV also demonstrates over a 70\% probability of reconstructing values close to the original. This highlights a significant privacy threat to VFL systems.}

\subsubsection{\textcolor{black}{Performance on Image Datasets}}

\begin{table}[tb]
\centering
\resizebox{0.45\textwidth}{!}{
\begin{tabular}
{lcccc}
\toprule
     & \textbf{UIFV-QA}   & \textbf{UIFV-DPA}         & \textbf{UIFV-IQA}          & \textbf{UIFV-SA}    \\ 
\midrule
PSNR & $23.83$        & $25.61$        & $14.92$        & $22.32$       \\
SSIM & $0.85$       & $0.89$         & $0.58$         & $0.81$         \\
\bottomrule
\end{tabular}
}
\caption{Reconstruction Performance of UIFV on the CIFAR10 dataset Across Four Scenarios.
}
\label{tab:image}
\end{table}

\begin{figure}[!h]
\centerline{\includegraphics[width=0.35\linewidth]{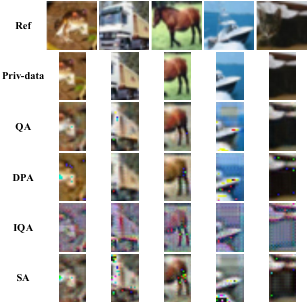}}
\caption{Our method is applied to the CIFAR10 dataset. The first line is the original image, the second and second lines are the private data that needs to be reconstructed during the VFL process, and the last four lines are the reconstruction effects under the four scenarios.}
\label{fig:cifar10}
\end{figure}

\textcolor{black}{To further demonstrate the generalizability of the UIFV method, we conducted reconstruction experiments on the CIFAR-10 image dataset, using PSNR and SSIM as metrics to evaluate the effectiveness of UIFV. The experimental results are shown in Table \ref{tab:image}. The results indicate that UIFV performs slightly better in the DPA scenario, which we attribute to the high spatial correlation in image data. This correlation benefits InverNet during shadow model training by facilitating the capture of relationships between intermediate features and the original image. Overall, UIFV exhibited strong attack performance across all four scenarios, posing a significant threat to VFL security. To provide a more intuitive understanding of our attack results, we present some actual reconstructed images in Figure \ref{fig:cifar10}.}

\begin{figure}[!h]
\centerline{\includegraphics[width=0.35\linewidth]{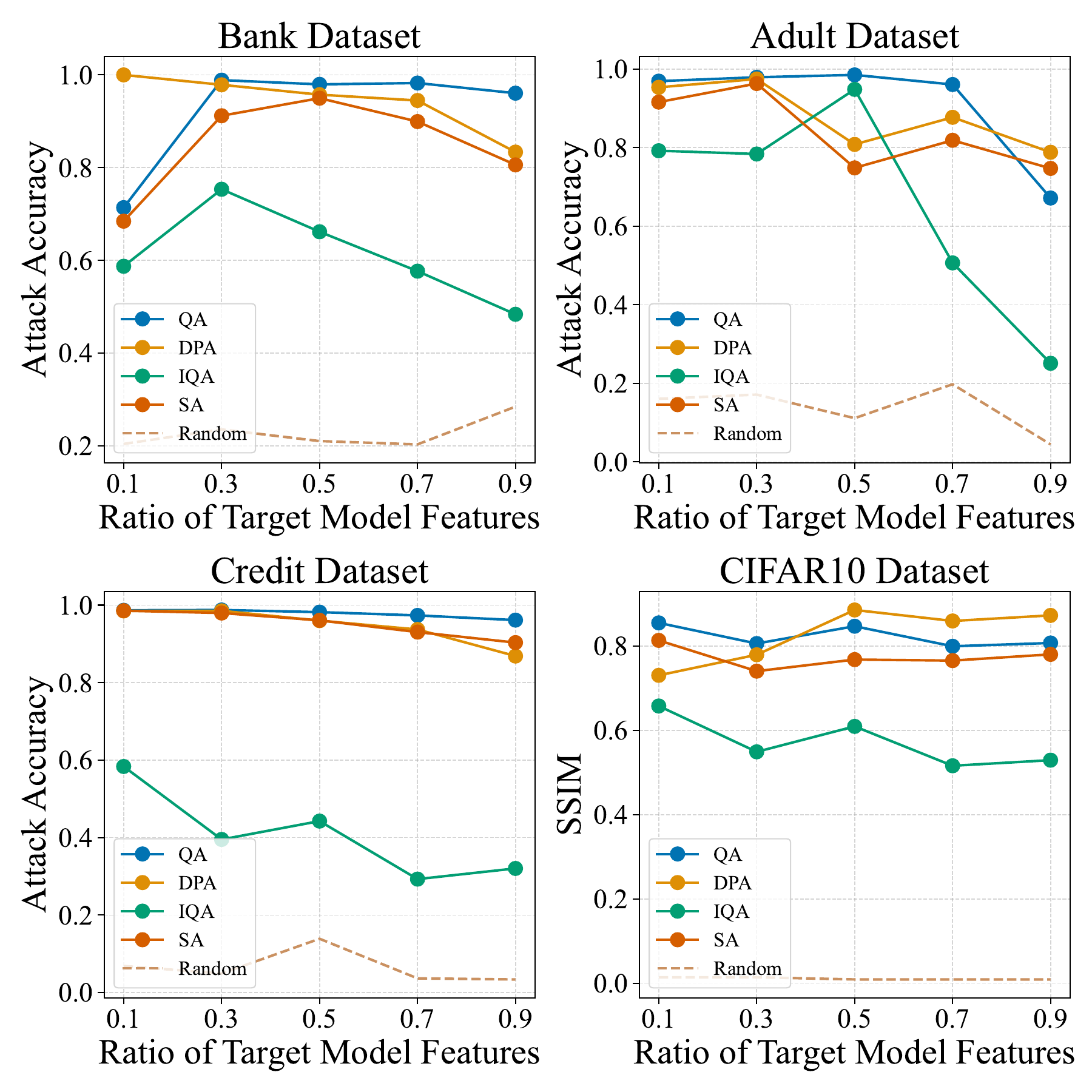}}
\caption{The attack accuracy of the target model with different ratios of features.}
\label{fig:RATIO}
\end{figure}

\subsubsection{Attack Effectiveness at Different Feature Splitting Ratios}

To explore realistic VFL scenarios with multiple parties holding different feature proportions, we tested various scenarios where the target party's feature proportion varied. We simulated five scenarios with two participants, where the target party's feature proportions were 0.1, 0.3, 0.5, 0.7, and 0.9, representing a range from highly imbalanced to balanced feature splitting. We evaluated our methods across four datasets, comparing them with random guessing.

The results, depicted in Fig. \ref{fig:RATIO}, show that different methods performed variably across feature splitting ratios and datasets. Generally, QA, DPA, and SA methods yielded stable and effective results, with attack accuracy rates over 60\%, indicating significant privacy risks. The IQA method was less stable but still outperformed random guessing.

\subsection{Defense evaluation}

\begin{table*}[htbp]
\centering
% \sisetup{
%     round-mode = places, % Rounds numbers
%     round-precision = 2, % to 2 decimal places
% }
\resizebox{0.5\textwidth}{!}{
\begin{tabular}{
  S[table-format=1.3]  % ratio列，最多一个整数位和三个小数位
  S[table-format=1.3]  % auc列，最多两个整数位和两个小数位
  S[table-format=2.2]  % m+d列
  S[table-format=2.2]  % d列
  S[table-format=2.2]  % m列
  S[table-format=2.2]|  % nothing列
  S[table-format=1.3]  % ratio列
  S[table-format=1.3]  % auc列
  S[table-format=2.2]  % m+d列
  S[table-format=2.2]  % d列
  S[table-format=2.2]  % m列
  S[table-format=2.2]  % nothing列
}
\toprule
 \multicolumn{6}{c|}{Bank} & \multicolumn{6}{c}{Income} \\
% \cmidrule(lr){1-6} \cmidrule(lr){7-12}
{Ratio} & {AUC} & {QA} & {DPA} & {IQA} & {SA} & {Ratio} & {AUC} & {QA} & {DPA} & {IQA} & {SA} \\
\midrule
1     & 0.859 & 86.38 & 88.29 & 18.96 & 18.96 & 1 & 0.888 & 87.78 & 84.92 & 73.92 & 73.87 \\
0.5   & 0.938 & 97.46 & 95.40 & 52.80 & 18.96 & 0.5 & 0.877 & 92.27 & 87.59 & 84.15 & 83.07 \\
0.1   & 0.938 & 97.93 & 95.50 & 64.01 & 78.16 & 0.1 & 0.876 & 93.51 & 89.29 & 82.91 & 81.45 \\
0.01  & 0.938 & 98.31 & 96.02 & 69.68 & 89.48 & 0.01 & 0.870 & 95.44 & 88.56 & 82.94 & 87.50 \\
0.001 & 0.939 & 98.22 & 96.04 & 66.40 & 90.49 & 0.001 & 0.867 & 95.86 & 89.15 & 61.72 & 85.48 \\
\midrule
\multicolumn{6}{c|}{Credit} & \multicolumn{6}{c}{CIFAR10} \\
% \cmidrule(lr){1-6} \cmidrule(lr){7-12}
{Ratio} & {AUC} & {QA} & {DPA} & {IQA} & {SA} & {Ratio} & {ACC} &{QA} & {DPA} & {IQA} & {SA} \\
\midrule
1     & 0.770 & 96.75 & 92.87 & 28.92 & 88.46 & 1     & 61.99 & 0.70 & 0.85 & 0.34 & 0.11 \\
0.5   & 0.760 & 96.52 & 96.52 & 26.04 & 90.45 & 0.5   & 67.26 & 0.14 & 0.81 & 0.14 & 0.0  \\
0.1   & 0.774 & 97.95 & 96.00 & 39.92 & 93.13 & 0.1   & 72.99 & 0.72 & 0.87 & 0.44 & 0.68  \\
0.01  & 0.771 & 97.59 & 96.25 & 41.50 & 93.26 & 0.01  & 74.84 & 0.80 & 0.88 & 0.53 & 0.74  \\
0.001 & 0.765 & 98.36 & 95.09 & 41.05 & 93.29 & 0.001 & 74.25 & 0.86 & 0.89 & 0.57 & 0.81  \\

\bottomrule
\end{tabular}
}
\caption{Experimental Results of DP Defense. For each dataset, the first column represents the ratio of the defense, the second column shows the results of the VFL task, and the last four columns indicate the effectiveness of our attack method under four different scenarios. For CIFAR10, SSIM is the Evaluation Metric for the Last Four Columns.}
\label{tab:dpdef}
\end{table*}

\begin{table*}[]
\centering
\resizebox{0.5\textwidth}{!}{
\begin{tabular}{
  S[table-format=1.3]  % ratio列，最多一个整数位和三个小数位
  S[table-format=1.3]  % auc列，最多两个整数位和两个小数位
  S[table-format=2.2]  % m+d列
  S[table-format=2.2]  % d列
  S[table-format=2.2]  % m列
  S[table-format=2.2]|  % nothing列
  S[table-format=1.3]  % ratio列
  S[table-format=1.3]  % auc列
  S[table-format=2.2]  % m+d列
  S[table-format=2.2]  % d列
  S[table-format=2.2]  % m列
  S[table-format=2.2]  % nothing列
}
\toprule
\multicolumn{6}{c|}{Bank} & \multicolumn{6}{c}{Income}\\
 % &  & {bank} & & & &  &  & {adult}& & & \\
 % \cmidrule(lr){1-6} \cmidrule(lr){7-12}
{Ratio} & {AUC} & {QA} & {DPA} & {IQA} & {SA} & {Ratio} & {AUC} & {QA} & {DPA} & {IQA} & {SA} \\
\midrule
1     & 0.937 & 18.96 & 89.19 & 49.02 & 18.96 & 1     & 0.885 & 70.87 & 69.93 & 56.21 & 3.07 \\
0.5   & 0.939 & 98.52 & 94.51 & 64.81 & 18.96 & 0.5   & 0.892 & 91.20 & 71.88 & 90.92 & 3.07 \\
0.1   & 0.940 & 98.22 & 95.48 & 67.16 & 90.67 & 0.1   & 0.870 & 96.37 & 90.68 & 73.40 & 87.19 \\
0.01  & 0.940 & 97.85 & 95.59 & 63.60 & 88.81 & 0.01  & 0.886 & 91.14 & 82.20 & 72.69 & 83.50 \\
0.001 & 0.939 & 98.26 & 96.20 & 60.42 & 91.19 & 0.001 & 0.868 & 94.90 & 89.17 & 78.52 & 86.16 \\
\midrule
% \multicolumn{6}{c}{credit} & \multicolumn{6}{c}{cifar10}\\
 \multicolumn{6}{c|}{Credit} & \multicolumn{6}{c}{CIFAR10} \\
{Ratio} & {AUC} & {QA} & {DPA} & {IQA} & {SA} & {Ratio} & {ACC} &{QA} & {DPA} & {IQA} & {SA} \\
\midrule
1     & 0.769 & 98.13 & 87.77 & 62.98 & 3.46  & 1     & 64.97 & 0.75 & 0.79 & 0.59 & 0.0 \\
0.5   & 0.775 & 98.35 & 96.08 & 57.51 & 94.34 & 0.5   & 66.81 & 0.88  & 0.79 & 0.62 & 0.07 \\
0.1   & 0.769 & 98.04 & 96.20 & 33.98 & 91.86 & 0.1   & 74.41 & 0.86  & 0.83 & 0.59 & 0.82 \\
0.01  & 0.772 & 97.89 & 96.24 & 37.66 & 91.96 & 0.01  & 74.13 & 0.81  & 0.87 & 0.56 & 0.76 \\
0.001 & 0.768 & 97.54 & 95.35 & 33.91 & 92.52 & 0.001 & 74.38 & 0.80 & 0.90  & 0.57 & 0.90 \\

\bottomrule
\end{tabular}
}
\caption{Experimental Results of Gaussian noise Defense.}
\label{tab:isodef}
\end{table*}

\textcolor{black}{Although this study primarily focuses on attack strategies, we have also examined several defensive measures. We first considered two common defensive measures: Differential Privacy (DP) and noise addition.} Differential Privacy technology defends against data reconstruction attacks by adding noise to gradients during the training process, thereby protecting individual data privacy while maintaining the overall effectiveness of the model. The method of adding noise to model outputs hinders attackers from obtaining precise information, thus protecting the data from malicious use, although this may affect the accuracy of the model.

To test the effectiveness of these defensive methods, we set the noise ratio to 1, 0.5, 0.1, 0.01, and 0.001, respectively, and conducted experiments across four different scenarios in four datasets. The experimental results for Differential Privacy are shown in Table \ref{tab:dpdef}, and those for noise addition are shown in Table \ref{tab:isodef}. These two defensive methods indeed reduce the effectiveness of data reconstruction attacks to some extent. However, their impact on attack effectiveness is relatively limited. In scenarios such as QA, DPA, and IQA, attacks can still maintain a certain success rate even with a high noise ratio. In the SA scenario, the impact of these two defensive methods is more noticeable. As the noise ratio increases, there is a downward trend in attack accuracy in the SA scenario.

\textcolor{black}{To further investigate the effectiveness of defense methods, we implemented two approaches designed to address privacy leakage in Federated Learning (FL). The first method is PA-iMFL\cite{wang2024pa}, a privacy amplification approach targeting data reconstruction attacks in advanced multi-layer federated learning. By combining local differential privacy, privacy-enhanced subsampling, and gradient sign resetting, PA-iMFL achieves bidirectional gradient compression, which not only improves communication efficiency but also strengthens privacy protection. The second method is VFLDefender\cite{zhu2024vulnerabilities}, which protects privacy by disrupting the correlation between gradients and training samples during model updates, thereby reducing attackers' ability to reconstruct labels or features.}

\begin{table}[ht]
\centering
\resizebox{0.39\textwidth}{!}{
\begin{tabular}{cccccc}
\toprule
Dataset & AUC/ACC & QA    & DPA   & IQA   & SA    \\
\midrule
Bank    & 0.925   & 98.65 & 92.98 & 80.17 & 94.08 \\
Income   & 0.893   & 89.55 & 77.24 & 73.82 & 58.32 \\
Credit  & 0.767   & 98.50 & 92.72 & 70.73 & 96.93 \\
CIFAR10 & 68.64   & 0.640 & 0.896 & 0.220 & 0.218 \\

\bottomrule
\end{tabular}
}
\caption{\textcolor{black}{Experimental Results of PA-iMFL Defense. For each dataset, the first column shows the results of the VFL task, and the last four columns indicate the effectiveness of our attack method under four different scenarios. For CIFAR10, SSIM is the Evaluation Metric.}} \label{tab:PAiMFL}
\end{table}

\begin{table}[ht]
\centering
\resizebox{0.39\textwidth}{!}{
\begin{tabular}{cccccc}
\toprule
Dataset & AUC/ACC & QA    & DPA   & IQA   & SA    \\
\midrule
Bank    & 0.865    & 98.67 & 93.04 & 80.60 & 89.11 \\
Income   & 0.891    & 78.53 & 75.86 & 77.27 & 83.72 \\
Credit  & 0.767    & 98.63 & 94.24 & 65.80 & 97.00 \\
CIFAR10 & 67.01   & 0.226 & 0.865 & 0.148 & 0.203 \\
\bottomrule
\end{tabular}
}
\caption{\textcolor{black}{Experimental Results of VFLDefender Defense. }
} \label{tab:VFLDefender}
\end{table}

\textcolor{black}{We evaluated the defensive effects of PA-iMFL and VFLDefender on four datasets. The experimental results for PA-iMFL are shown in Table \ref{tab:PAiMFL}, while those for VFLDefender are shown in Table \ref{tab:VFLDefender}. The results demonstrate that both defense methods have a minimal impact on the primary VFL training tasks and effectively reduce the efficacy of data reconstruction attacks to some extent, particularly showing significant defensive effects on image datasets. However, neither method provided effective defense in the DPA scenario. This is primarily because, in the DPA scenario, the attack does not require any query requests to the target party, rendering communication-focused defense measures partially ineffective.}

\subsection{Ablation Study}

\subsubsection{Size of the auxiliary dataset}

\begin{figure}[]
\centerline{\includegraphics[width=0.38\linewidth]{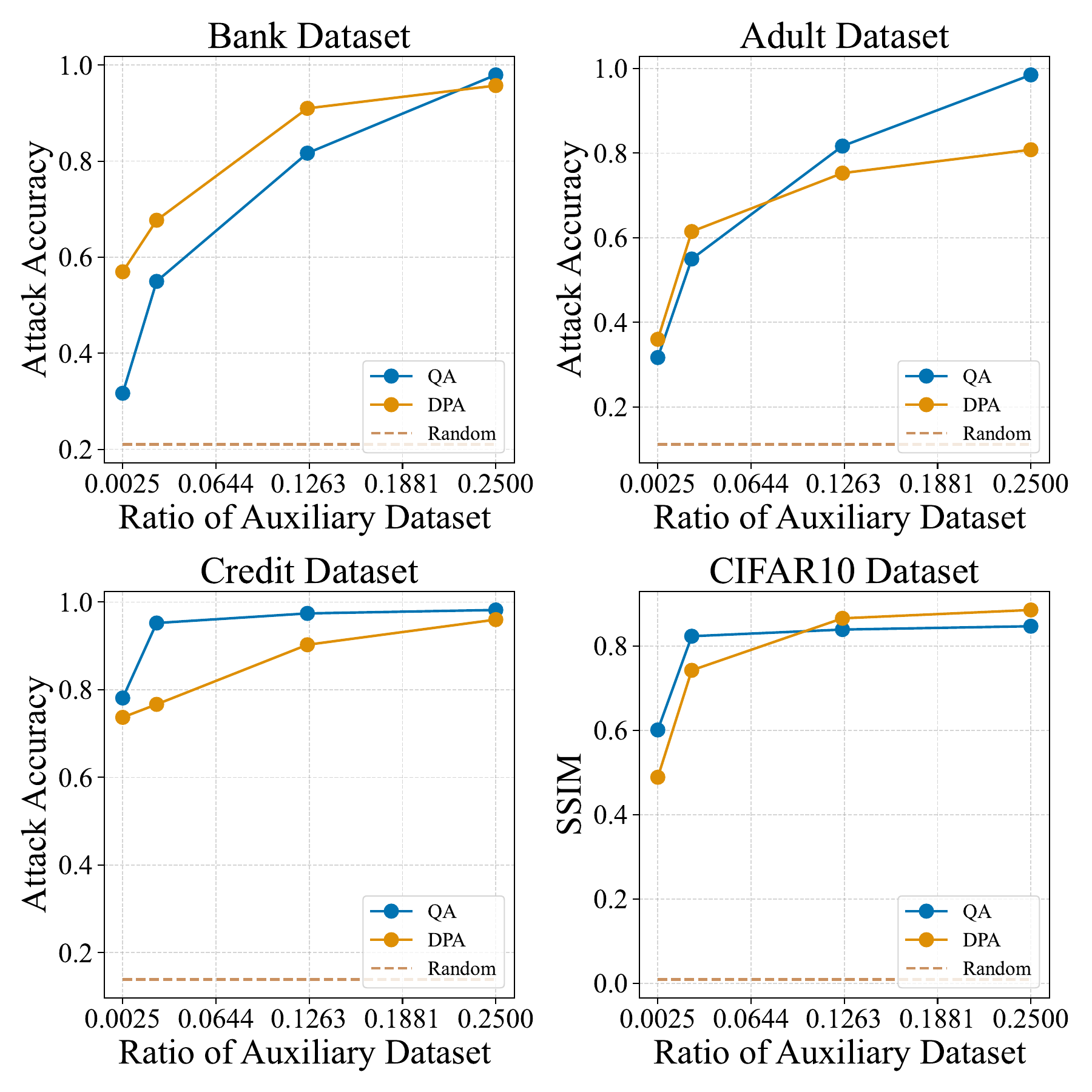}}
\caption{The size ratio of the auxiliary dataset relative to the training dataset.}

\label{fig:auxiliary_RATIO}
\end{figure}

Among the four attack scenarios, QA and DPA rely on i.i.d. auxiliary datasets for data reconstruction. Our study found a direct correlation between the auxiliary dataset size and reconstruction accuracy.

We tested four sizes of auxiliary datasets: 0.0025, 0.025, 0.125, and 0.25, representing their relative sizes to the VFL training dataset. Results in Fig. \ref{fig:auxiliary_RATIO} show that as the auxiliary dataset size increases, the reconstruction accuracy of our method improves. This suggests that a larger auxiliary dataset, offering more information, allows for a more accurate estimation of the original dataset's distribution, thus enhancing reconstruction accuracy.

\subsubsection{Impact of Data Generation}\label{sec:DG}

\begin{table}[htbp]
\centering
% \begin{tabular}{@{}llll@{}}
\resizebox{0.45\textwidth}{!}{
\begin{tabular}{lccc}
\toprule
\textbf{Dataset} & \textbf{Evaluation}  & \textbf{IQA-No-DG  } & \textbf{IQA}           \\ 
\midrule
\multirow{3}{*}{Bank} & Accuracy         & $27.19 \pm 5.27$       & \bm{$66.17 \pm 1.43$}      \\
              & Discrete Acc     & $42.30 \pm 9.59$       & \bm{$87.78 \pm 2.11$   }   \\
              & Continuous Acc   & $12.08 \pm 1.19$       & \bm{$44.57 \pm 1.21$  }    \\
\multirow{3}{*}{Income} & Accuracy         & $53.09 \pm 3.59$       & \bm{$94.81 \pm 0.54$  }    \\
              & Discrete Acc     & $57.88 \pm 4.64$       & \bm{$98.08 \pm 0.57$  }    \\
              & Continuous Acc   & $33.95 \pm 2.42$       & \bm{$81.72 \pm 0.71$  }    \\
\multirow{3}{*}{Credit} & Accuracy         & $14.35 \pm 2.56$       & \bm{$44.25 \pm 1.61$ }     \\
              & Discrete Acc     & $36.94 \pm 7.06$       & \bm{$75.17 \pm 4.78$   }   \\
              & Continuous Acc   & $1.45 \pm 0.26$        & \bm{$26.59 \pm 1.84$   }   \\ 
\multirow{2}{*}{CIFAR10} & PSNR         & $14.26 \pm 0.16$       & \bm{$14.93 \pm 0.49$ }     \\
              & SSIM     & $0.56 \pm 0.01$       & \bm{$0.58 \pm 0.04$ }     \\
\bottomrule
\end{tabular}
}
\caption{Comparative Performance of IQA with and without Data Generation Module Across Different Datasets. (Best
results are highlighted in bold.)}
\label{tab:DG}
\end{table}
In the IQA attack scenario, we utilize a Data Generator (DG) to enhance the accuracy of reconstruction attacks. To assess the DG module's effectiveness, we compared it with a random number generator simulating a uniform distribution. Results in Table \ref{tab:DG} show that IQA with the DG module significantly improved performance on all tabular datasets and metrics, averaging a 30\% increase in accuracy, with notable gains in image datasets. This demonstrates the DG module's vital role in simulating the original dataset's distribution and increasing attack accuracy, in contrast to the lower performance with a simple random number generator. Hence, the DG module is essential for successful data reconstruction attacks.

\subsubsection{Size of the Known Private Dataset}

\begin{figure}[htbp]
\centerline{\includegraphics[width=0.5\linewidth]{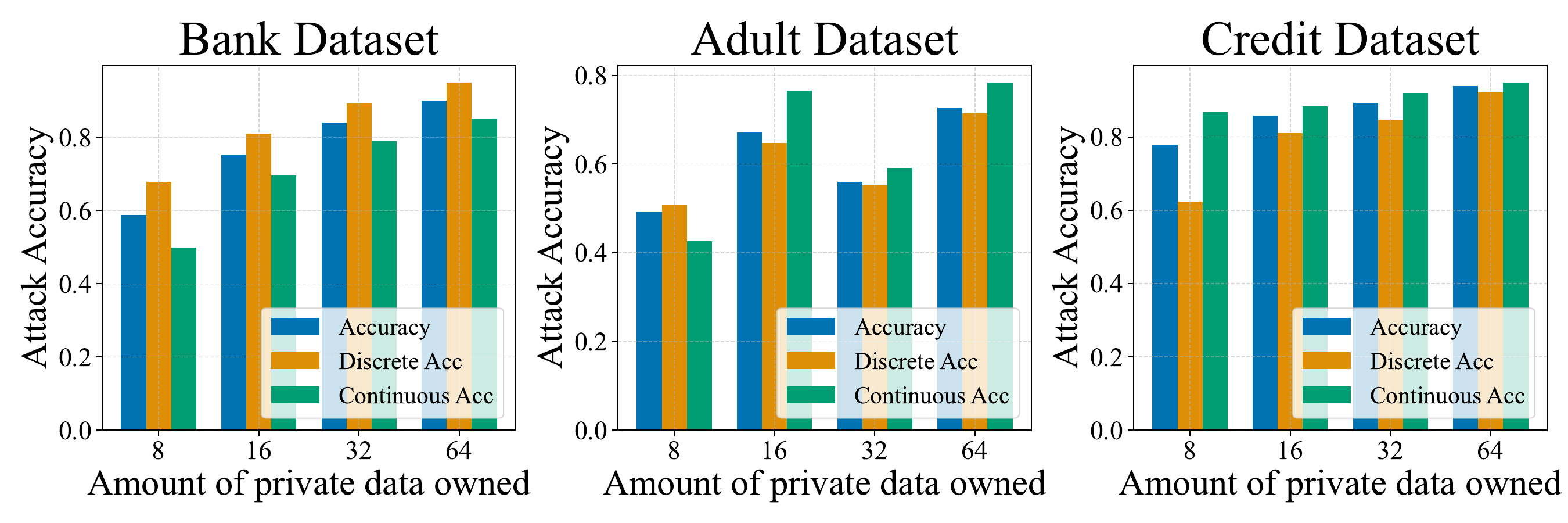}}
\caption{The relationship between the amount of private data owned and attack accuracy on the Bank, adult and Credit datasets.}
\label{fig:SA}
\end{figure}

In the SA attack scenario, we assume that the attacker has acquired a small number of the target's private data to train an InverNet for data reconstruction. We evaluated the attack's efficacy with varying numbers of prior-known private samples, 8, 16, 32, and 64, as shown in Fig. \ref{fig:SA} and Table \ref{tab:SA-image}.

The results indicate a positive correlation between the number of training samples and the accuracy metrics (general, categorical, continuous, PSNR, and SSIM). This demonstrates that even with a minimal amount of training data, such as 8 samples, our method can still be successful and yield reasonable results.

\begin{table}[!tbp]
\centering
\sisetup{
  round-mode = places, % Rounds numbers
  round-precision = 2, % to 2 places
}
\resizebox{0.35\textwidth}{!}{
\begin{tabular}{
  c
  S[table-format=2.2]
  S[table-format=2.2]
  S[table-format=2.2]
  S[table-format=2.2]
}
\toprule
% \diagbox{\textbf{Eval}}{\textbf{$x^{priv}$ num}}
\textbf{$\bm{x^{priv}}$ num.} & \bm{$8$} & \bm{$16$} & \bm{$32$} & \bm{$64$} \\ 
\midrule
PSNR & 19.3122099 & 21.00210019 & 21.58687616 & 22.31927847 \\ 

SSIM & 0.737310911 & 0.778440135 & 0.790168558 & 0.81262031 \\ 
\bottomrule
\end{tabular}
}
\caption{The relationship between the amount of private data owned on the CIFAR10 dataset and the reconstruction effect.}
\label{tab:SA-image}
\end{table}

\subsubsection{The Impact of InverNet Model Size on Attack Effectiveness}

As a crucial component of the UIFV framework, we delved into the impact of the size of the InverNet model on our attack performance. For attacks on tabular data, we employed InverNet, which is composed of three fully connected layers. During the experiments, we assessed the specific effects of one and two fully connected layers on the effectiveness of the attack. Similarly, for attacks on image data, we used InverNet constructed with two layers of transposed convolution, and evaluated the impact of one and three layers of transposed convolution on the results of the attack.

\begin{table*}[htbp]
\centering
\resizebox{0.5\textwidth}{!}{
\begin{tabular}{
  S[table-format=1.0]  % layer列
  S[table-format=2.2]  % m+d列
  S[table-format=2.2]  % d列
  S[table-format=2.2]  % m列
  S[table-format=2.2]| % no列
  S[table-format=1.0]  % layer列
  S[table-format=2.2]  % m+d列
  S[table-format=2.2]  % d列
  S[table-format=2.2]  % m列
  S[table-format=2.2]  % no列
}
\toprule
 \multicolumn{5}{c|}{Bank} & \multicolumn{5}{c}{Income} \\
{Layers} & {QA} & {DPA} & {IQA} & {SA} & {Layers} & {QA} & {DPA} & {IQA} & {SA} \\
\midrule
1 & 95.39 & 93.51 & 58.09 & 89.12 & 1 & 92.45 & 86.67 & 76.08 & 86.50 \\
2 & 97.66 & 95.95 & 59.85 & 89.30 & 2 & 93.33 & 86.23 & 75.10 & 86.44 \\
3 & 97.96 & 95.74 & 66.17 & 90.07 & 3 & 98.49 & 80.80 & 94.81 & 72.79 \\
\midrule
\multicolumn{5}{c|}{Credit} & \multicolumn{5}{c}{CIFAR10} \\
{Layers} & {QA} & {DPA} & {IQA} & {SA} & {Layers} & {QA} & {DPA} & {IQA} & {SA} \\
\midrule
1 & 97.73 & 94.01 & 40.40 & 90.30 & 1 & 0.70 & 0.76 & 0.54 & 0.70 \\
2 & 98.15 & 95.93 & 40.89 & 93.56 & 2 & 0.85 & 0.89 & 0.61 & 0.81 \\
3 & 98.19 & 95.99 & 44.25 & 93.83 & 3 & 0.86 & 0.90 & 0.62 & 0.83 \\
\bottomrule
\end{tabular}
}
\caption{The Impact of InverNet Model Size on Attack Effectiveness. For each dataset, the first column represents the number of layers of InverNet, and the last four columns indicate the effectiveness of our attack method under four different scenarios. For CIFAR10, SSIM is the Evaluation Metric for the Last Four Columns.}
\label{tab:model-size}
\end{table*}

The experimental results are shown in Table \ref{tab:model-size}. The results showed that as the number of layers and the size of the InverNet model increased, there was a certain degree of enhancement in the performance of UIFV attacks. However, overall, while increasing the number of layers in InverNet does improve performance, once it exceeds a certain threshold, the growth in performance tends to saturate. When applying the UIFV framework, choosing the appropriate size of the InverNet model is particularly important, necessitating careful adjustment and selection based on different data types and attack scenarios.

\section{Conclusions}\label{sec:7}

\subsection{Summary}

In our paper, we introduced the Unified InverNet Framework in VFL (UIFV), a novel approach for conducting data reconstruction attacks in VFL environments. Unlike traditional attack strategies, UIFV leverages intermediate features of the target model rather than relying on gradient information or model parameters. UIFV exhibits remarkable adaptability and is effective across various black-box scenarios. Experiments conducted on four benchmark datasets show that our approach surpasses the existing attack methods in effectiveness, achieving over $96\%$ accuracy in scenarios like QA. Through comprehensive ablation studies, we also confirmed the importance of key components, such as the data generation module. \textcolor{black}{Our research expands the understanding of VFL data reconstruction attacks and provides new insights for privacy protection. The UIFV framework showcases its high applicability and precision across multiple scenarios, offering practical guidance for designing more robust defense mechanisms in the future. Additionally, our findings highlight privacy vulnerabilities in VFL systems under real-world applications, providing valuable support for policy-making and technological advancements in data protection.}

\subsection{\textcolor{black}{Limitation}}
\textcolor{black}{Despite its strong experimental performance, the UIFV framework has certain limitations. First, this study primarily focuses on scenarios where the attacker is an active participant in the VFL system. However, when the attacker acts as a passive participant, the conditions for a successful attack become significantly more stringent. For instance, in Data Passive Attack (DPA) scenarios, passive participants may find it challenging to train effective shadow models, potentially limiting the applicability of the UIFV method. Furthermore, our research is based on the most generic VFL architectures. More complex VFL setups could introduce additional challenges for the UIFV framework.}

\subsection{\textcolor{black}{Future work}}

\textcolor{black}{We consider adapting the UIFV framework to accommodate more complex VFL architectures and extend its application to real-world scenarios in healthcare, finance, and the Internet of Things (IoT). These efforts will help validate the framework’s broad applicability and practical impact. Additionally, considering the privacy risks exposed by the UIFV framework, future research should focus on developing more advanced defense mechanisms, such as purification defense strategies \cite{yang2023purifier}, to comprehensively enhance the security of VFL systems.}

\bibliographystyle{elsarticle-num} 
\bibliography{refs}




% %% For numbered reference style

%   \textit{\LaTeX: a document preparation system},
%   Addison Wesley, Massachusetts,

% \end{thebibliography}
\end{document}